# Bayesian Distributional Models of Executive Functioning

Robert Kasumba[1], Zeyu Lu[2], Dom CP Marticorena[3], Mingyang Zhong[3], Paul Beggs[3], Anja Pahor[4], Geetha Ramani[5], Imani Goffney[6], Susanne M Jaeggi[7], Aaron R Seitz[7], Jacob R Gardner[8], Dennis L Barbour[1,2,3]

[1]Division of Computing and Data Science, Washington University, 1 Brookings Drive, St. Louis, MO 63130
[2]Department of Computer Science and Engineering, Washington University, 1 Brookings Drive, St. Louis, MO 63130
[3]Department of Biomedical Engineering, Washington University, 1 Brookings Drive, St. Louis, MO 63130
[4]Department of Psychology, University of Maribor, 2000 Maribor, Slovenia
[5]Department of Human Development and Quantitative Methodology, University of Maryland, College Park, MD 20742-5025
[6]Department of Teaching and Learning, Policy and Leadership, University of Maryland, College Park, MD 20742-5025
[7]Department of Psychology, Northeastern University, Boston, MA 02115
[8]Department of Computer and Information Science, University of Pennsylvania, Philadelphia, PA 19104

Corresponding Author:
Dennis L Barbour
1 Brookings Drive
MSC 1097-202-190
St. Louis, MO 63130
dbarbour@wustl.edu

# Abstract

This study uses controlled simulations with known ground-truth parameters to evaluate how Distributional Latent Variable Models (DLVM) and Bayesian Distributional Active LEarning (DALE) perform in comparison to conventional Independent Maximum Likelihood Estimation (IMLE). DLVM integrates observations across multiple executive function tasks and individuals, allowing parameter estimation even under sparse or incomplete data conditions. To establish known-ground truth, we uniformly sample individual sessions from a neural network learned latent space and map them to distributional cognitive performance across different tasks. The individual test-items are then sampled from these distributions using either DALE, random procedure or a standard fixed battery approach. When given the same set of observations, DLVM consistently outperformed IMLE, especially under smaller amounts of data, and converges faster to highly accurate estimates of the true distributions. In a second set of analyses, DALE adaptively guided sampling to maximize information gain, outperforming random sampling and fixed test batteries, particularly within the first 80 trials. These findings establish the advantages of combining DLVM's cross-task inference with DALE's optimal adaptive sampling, providing a principled basis for more efficient cognitive assessments.

# Introduction

Executive functioning (EF) is fundamental to an individual's ability to carry out daily activities, encompassing essential cognitive processes such as problem-solving, decision-making and goal-directed behavior. EF is generally understood to comprise three core constructs: working memory, cognitive flexibility and inhibitory control (Friedman & Miyake, 2017; Miyake et al., 2000). These constructs govern our ability to process and retain information, regulate responses and adapt to changing environments, playing a crucial role in key life outcomes such as academic achievement, career success and overall well-being (Diamond, 2013).

Understanding an individual's EFs is of significant interest in psychology and cognitive science as it could enable the development of targeted interventions. For example, assessing EFs for individual students can inform optimal study strategies for them, particularly in subjects requiring high cognitive demand such as mathematics (Diamond, 2013). Cognitive assessment typically relies on a battery of standardized tests designed to estimate specific EF constructs. The Corsi Span task (Kessels et al., 2008), for instance, is a spatial recall task commonly used to evaluate working memory, while the Stroop reading/color suppression task is employed to assess inhibitory control (Gutiérrez-Martínez et al., 2018; Linkovski et al., 2021).

Traditional approaches to estimating EFs assume that cognitive tasks provide construct-specific information and/or that there are linear relationships between performance on tasks. Given the noisy nature of the data from behavioral tasks, methods assuming independence typically involve administering repeated identical trials and deriving an estimate by averaging across them. For example, the mean response time is used to quantify Stroop effect performance reflecting inhibitory control. Advanced methods such as Structural Equation Models (SEM), which aim to infer relationships across cognitive constructs, typically assume linear relationships, potentially oversimplifying the true underlying interactions (Friedman & Miyake, 2017). These approaches collectively treat trial-to-trial variability as noise rather than capturing meaningful intra-individual variability (Hedge et al., 2018; Rouder & Haaf, 2019).

To provide a more general behavioral modeling framework, we recently introduced Distributional Latent Variable Modeling (DLVM), a nonlinear machine learning framework that simultaneously extracts cross-construct information from multiple cognitive tasks using item-level observations (Kasumba et al., 2025). In this framework, each observation contributes to the estimation of multiple constructs through the learned latent embeddings, enhancing both efficiency and predictive accuracy. DLVM also enables estimation of cognitive performance on tasks with limited or missing task data by leveraging information from other administered tasks to perform construct-level inference.

In the same work, we also introduced Distributional Active LEarning (DALE), a Bayesian method for adaptive task and item selection. DALE identifies the most informative cognitive task items to administer to each individual, thereby customizing the testing process for each session. Together, DLVM and DALE (DLVM+DALE) offer a principled framework for efficient and individualized cognitive assessment. We demonstrated that DLVM+DALE EF estimates converge substantially faster than traditional methods that assume task independence, requiring only one-third the number of trials under the conditions tested.

In the present study, we expand upon these findings by evaluating the performance of DLVM and DALE under conditions where the ground truth is known. In Kasumba et al. (2025), we assessed convergence using data acquired from actual participants with real-time Bayesian optimization. While ecologically valid, that approach did not allow for precise evaluation of estimation accuracy, as ground truth model parameters were unknown for these individuals. Here, we address this limitation using simulation. We generate trial-level responses from known cognitive performance parameters derived from actual individuals. This simulation oracle allows us to flexibly generate as many trials as needed, circumventing the limitations of human data collection.

We conduct two sets of comparisons. First, we evaluate the accuracy of DLVM in estimating latent cognitive parameters as a function of data quantity relative to Independent Maximum Likelihood Estimation (IMLE), which provides optimal estimates assuming task independence. Second, we examine how different data sampling (i.e., task item selection) strategies affect estimation accuracy. Specifically, we compare 1) DLVM + DALE, 2) DLVM + random task sampling and 3) a traditional strategy delivering task items in a sequential block design.

We test two key hypotheses:

**H1:** DLVM provides more accurate estimates than IMLE when data are sparse, but the more flexible IMLE overtakes DLVM when sufficient data are available.

**H2:** DLVM+DALE, with its ability to identify and prioritize informative tasks, provides more accurate estimates than both DLVM + random sampling and traditional IMLE.

# Background

## Executive functioning

Executive functioning (EF) consists of three primary constructs: working memory, inhibitory control and cognitive flexibility. Working memory involves the ability to hold and manipulate information temporarily. Inhibitory control enables an individual to suppress automatic responses and resist distractions. Cognitive flexibility refers to the ability to switch between tasks or perspectives when needed (Diamond, 2013; Miyake et al., 2000).

The development and efficacy of these EFs are crucial for success in academic, social and professional domains. In particular, EF performance has been linked to important cognitive outcomes, such as performance in mathematics (Cragg & Gilmore, 2014), where tasks often demand working memory and the ability to shift between strategies (Friedman & Miyake, 2017). Despite their importance, understanding how these constructs interact and develop over time is still a topic of ongoing research. For example, while EFs are known to mature into a relatively stable form by around the age of 10, much remains to be discovered about their continued development throughout the lifespan (Younger et al., 2023).

## Conventional approaches to modeling executive functioning

To assess these functions, cognitive testing often employs standardized, decontextualized tasks designed to target specific EF constructs. These tests, such as the Corsi Span (which assesses working memory) and the Stroop test (which evaluates inhibitory control), are commonly used in cognitive psychology (Brunetti et al., 2014; Scarpina & Tagini, 2017). However, conventional methods targeting single-construct performance typically assume that each test reflects predominantly the single construct it is designed to measure, largely ignoring potential interactions or shared variance between tasks that reflect different EF domains (Hedge et al., 2018). Moreover, these approaches tend to rely on summary statistics, such as mean response times, which do not fully account for individual differences or the variability inherent in cognitive performance across repeated trials (Rouder et al., 2005).

Structural Equation Models (SEM) are often employed to model cross-task relationships, assuming they are linear and have a pre-defined structure. This linearity assumption oversimplifies the relationships observed in cognitive test variables, however, and may fail to fully capture the complexity of individual differences (Friedman & Miyake, 2017). This shortcoming invites the use of more complex modeling approaches to capture potential non-linear and nuanced relationships better. More advanced statistical approaches, such as hierarchical Bayesian modeling (Rouder & Haaf, 2019; Veenman et al., 2024) and diffusion modeling (Ratcliff & McKoon, 2008; Ratcliff & Tuerlinckx, 2002), offer more nuanced ways of capturing individual differences in EF by considering latent dynamic parameters rather than simple mean-based measures, though still generally representing linear relationships.

## Distributional Latent Variable Models (DLVM)

The limitations of traditional modeling techniques described above prompted our development of the Distributional Latent Variable Model (DLVM) (Kasumba et al., 2025), a nonlinear machine learning framework that allows a more generalized representation of EFs. Unlike conventional approaches that focus mainly on averages, DLVM captures the complex, nonlinear interrelations among tasks and offers a richer understanding of performance variation both within and between individuals. These features make it particularly well-suited for modeling intra-individual variability as well as individual differences.

A key strength of DLVM is its ability to quantify uncertainty in cognitive assessments, thereby uncovering insights that traditional methods may miss. Specifically, DLVM jointly models both the central tendency (e.g., mean performance) and variability (e.g., fluctuations, standard deviations) of cognitive data across individuals and tasks. This dual perspective provides a more complete picture of performance and underlying dynamics. DLVM is also highly data efficient. By integrating Bayesian active machine learning strategies, it can identify and prioritize the most informative data points for each individual and testing session, allowing researchers to achieve predictive accuracy comparable to traditional methods while using only a fraction of the data.

At a technical level, DLVM works by compressing task performance into a low-dimensional latent structure. For instance, a DLVM model can capture nonlinear dependencies between test items, task-level performance and individual differences. Within this learned latent space, each individual occupies a unique position that reflects their cognitive profile. These latent representations can then be used to recover task-level performance or, more powerfully, to generate new instances of plausible cognitive behavior. In this way, a trained DLVM functions as a generative oracle, capable of producing task response data for an unlimited variety of realistic cognitive profiles for exploration and simulation.

# Materials and Methods

This work extends our previous research by evaluating the performance of DLVMs under conditions where the ground-truth parameters are known. In the prior work, we applied DLVM and DALE to data collected in real time from human participants and demonstrated comparable accuracy at the individual test level to traditional fixed test battery approaches. Here, we extend these findings using simulations, which allow us to precisely control and know the underlying cognitive parameters. By leveraging this ground-truth setting, we can rigorously assess the accuracy of model estimates and gain deeper insights into the strengths and limitations of DLVM and adaptive sampling strategies, beyond what is possible with human data alone.

## Model Training Dataset

A simulated dataset was first generated from trained DLVM models. To train the DLVM models, we used a retrospective dataset from our previous study, consisting of 18 participants who completed up to 10 cognitive test sessions over 10 days via a mobile app. The participants completed a full test battery in each of the 10 sessions. All participants provided informed consent under an IRB-approved protocol and received $10 per session. Sessions were self-directed and conducted remotely. As previously justified, we exclude 79 sessions due to missing task data stemming from technical issues and 13 additional sessions due to outlier performance (Kasumba et al., 2025). The final dataset includes 88 valid sessions, each treated as an independent observations without modeling within-participant correlations across days. We refer to this

training dataset as the COLL10 dataset. We provide additional analysis of model performance without removing outliers in the Supplementary Material.

## Cognitive Test Battery

The test battery in the COLL10 dataset comprises eight assessment tasks, including Paced Auditory Serial Addition Test (PASAT+), Countermanding, Running Span (with 2 and 3 items), Numerical Stroop, Magnitude Comparison, Corsi Simple Span, Corsi Complex Span and Cancellation (Pahor et al., 2019, 2021, 2022; Rojo et al., 2023). These tests target specific constructs of the primary executive functions, i.e., working memory, cognitive flexibility and inhibitory control. **Table 1** shows the constructs reflected by each task and the associated distributional parameters. Twelve distributional parameters in total characterize the performance of an individual for a session with this test battery.

## Model training Procedure

We trained a neural-network-based DLVM following the procedure described in detail previously (Kasumba et al., 2025). Specifically, we fit three different models of latent dimensionalities 1, 2 and 3 using the COLL10 dataset to learn the mapping from latent representations to the observation parameter space. The parameter space characterizes the distributional summary of an individual's performance across tasks. This process is illustrated for the current test battery in **Figure 1**.

We selected these latent dimensionalities to reflect theory-informed, plausible low-dimensional structures of executive functioning. Although executive functioning is often described in terms of three major components, such as working memory, cognitive flexibility, and inhibitory control (Friedman & Miyake, 2017; Miyake et al., 2000), , other work has also supported more unified or otherwise lower-dimensional accounts (Löffler et al., 2024; Younger et al., 2023). Accordingly, our simulations examined one-, two-, and three-dimensional latent spaces to evaluate model behavior across a plausible range of latent dimensionalities, rather than to impose a one-to-one correspondence between latent dimensions and specific EF constructs. The training process simultaneously estimated the individual-specific latent variables and their mapping to task-specific performance. Models were trained for up to 10,000 epochs with a learning rate of 0.01 and a loss penalty of 0.01.

Because this work aims to characterize the DLVM framework and its integration with active machine learning, rather than to optimize a single model specification, our analyses mostly focus on the 2-dimensional model. This choice reflects the ease of visualization of its latent variable space and correspondingly improved interpretability, while still providing reasonable fits and capturing the underlying patterns in the data. Additional results from the 1- and 3-dimensional models are reported in the Supplementary Material. We refer to these models as DLVM-1, DLVM-2 and DLVM-3, corresponding to 1-dimensional, 2-dimensional and 3-dimensional conditions.

## Data Simulation

We constructed a simulated dataset of cognitive performance across multiple tasks for evaluation in this study using trained DLVM models. Specifically, we defined a regular grid spanning the full range of each latent variable and randomly selected 88 grid points (to match the size of the training set). This procedure ensured systematic coverage of the latent space and had the added advantage of sampling well outside the distribution clusters of training sessions, thereby providing a natural evaluation of external validity. For each sampled point, the DLVM model was then used to map the latent representation to the corresponding distributional parameter space, producing the ground-truth parameters. This procedure guarantees a well-defined mapping between the latent representation and the joint cognitive performance profile of an individual across tasks. **Figure 2** illustrates the latent space with the training points, the sampled validation points and their mappings to the parameter space. As shown, correlations naturally emerge between different tasks. For example, sessions with slower response times on Stroop and Countermanding tasks tend to exhibit lower working memory thresholds (i.e., Simple and Complex Corsi $\psi_\theta$).

From these ground-truth distributional parameters, we generated eight task-specific distributions, summarized in **Table 1**. For each distribution, 240 trial-level observations were simulated, providing the data used by the different models to attempt recovery of the ground truth parameters.

## Bayesian Distributional Active LEarning (DALE)

To infer individual-level cognitive parameters from trial-level data, we employed the Bayesian DALE algorithm introduced and described in detail in Kasumba et al. (2025). DALE frames the problem of cognitive testing as a sequential Bayesian inference process, where each new trial is chosen adaptively to maximize the expected information gained about an individual's latent representation. Rather than administering a fixed battery of tasks, DALE dynamically personalizes the assessment process so that each participant receives a unique sequence of trials tailored to their inferred cognitive profile.

Bayesian active machine learning has been successfully applied in audition (Barbour et al., 2019; Cox & de Vries, 2021; Heisey et al., 2018, 2020; Schlittenlacher et al., 2018; Song et al., 2015; Twinomurinzi et al., 2024), vision (Chesley & Barbour, 2020; Marticorena, Wong, Browning, Wilbur, Davey, et al., 2024; Marticorena, Wong, Browning, Wilbur, Jayakumar, et al., 2024) and more general psychometric field estimation (Song et al., 2017, 2018). In those studies, Gaussian process models combined with active stimulus selection yielded rapid, individualized estimates of perceptual thresholds and psychometric functions while dramatically reducing the number of trials required. This prior work has demonstrated the value of treating psychophysical testing as a Bayesian optimization problem, where adaptively chosen stimuli efficiently resolve uncertainty about latent perceptual functions. DALE extends this paradigm beyond sensory domains to higher-order cognition by generalizing the Bayesian active learning framework to distributional models of cognitive performance. While these previous methods aimed to recover continuous perceptual functions such as audiograms, DALE focuses on estimating cognitive parameter distributions that characterize latent traits such as working memory, attentional control or decision variability. In doing so, DALE generalizes the principles of probabilistic classification and Bayesian active learning from the perceptual domain with distributional inference over individual differences, enabling scalable and precise cognitive assessment.

Briefly, the process begins by administering a random trial to a new individual. The outcome, denoted $y_1$, provides the first piece of evidence linking the unobserved latent representation $\mathbf{z}$ to the observation model defined by the DLVM. Given this outcome, DALE computes an initial posterior distribution:

$$p(\mathbf{z}|y_1) \propto p(y_1|\mathbf{z})p(\mathbf{z}),$$

where $\mathbf{z}$ is the prior over latent positions and $p( y_1 \mid \mathbf{z} )$ is the likelihood derived from the DLVM mapping. The posterior is approximated via gradient-based optimization using the Adam optimizer (Kingma & Ba, 2017), which identifies the latent configuration that maximizes posterior probability. This posterior distribution represents the evolving estimate of the individual's latent position, incorporating both central tendency and uncertainty.

At each subsequent iteration $t$, DALE selects the next trial $x_t$ using an information-theoretic acquisition rule. Specifically, it evaluates the mutual information between the latent representation $\mathbf{z}$ and the potential trial outcome $y_t$:

$$I(y_t; \mathbf{z}|D_{t-1}) = H(p(y_t|D_{t-1})) - E_{q(\mathbf{z})}[H(p(y_t|D_{t-1}, \mathbf{z}))],$$

where $D_{t-1}$ denotes the data observed up to the previous iteration and $q(\mathbf{z})$ denotes the variational approximation to the posterior $p(\mathbf{z}|D_{t-1})$. The candidate trial with the largest expected information gain is selected. Intuitively, this procedure ensures that each new trial is chosen to maximally reduce posterior entropy, thereby producing the greatest expected refinement of the latent estimate. Once the trial is administered and the outcome $y_t$ is observed, the posterior is updated sequentially:

$$p(\mathbf{z}\,|D_t) \propto p(y_t|\mathbf{z})p(\mathbf{z}|D_{t-1}).$$

The posterior from the previous iteration is used as the prior for the next. This recursive Bayesian updating maintains full uncertainty propagation across the testing sequence and guarantees that all past evidence informs subsequent inference.

For improved performance, DALE can also be primed with an initial batch of observations before active learning is turned on. This initialization ensures that every task contributes some evidence early in the procedure, for example, by sampling at least two observations per task. The priming data may be obtained via any sampling scheme and can be delivered either all at once or sequentially. In our implementation, we adopt the sequential procedure, updating the latent estimate $\mathbf{z}$ after each new observation. This strategy improves stability in the early stages of inference and reduces the risk of poor initialization.

The process continues until a predefined trial budget $T$ is exhausted, although DALE could also employ a stopping rule based on posterior precision (e.g., halting when entropy falls below a threshold (Song et al., 2015)). By design, the algorithm produces individualized trial sequences that reflect both the cognitive profile of each participant and the evolving uncertainty in its estimation.

## Independent Maximum Likelihood Estimation (IMLE)

The IMLE model assumes that cognitive tests are independent of one another (Cousineau et al., 2004; Embretson, 1991). For each session and each test, distributional parameters were computed using the maximum likelihood estimator applied to item-level observations. When a closed-form solution was available (e.g., lognormal distribution for reaction times or binomial distribution for accuracy-based tasks), it was used directly. For tasks modeled using sigmoid functions, such as span tasks where no closed-form solution exists, we employed gradient descent optimization to estimate the parameters that maximized the likelihood of the observed responses. Just like DLVM, the IMLE model outputs a performance vector consisting of 12 distributional parameters for each session.

## Analysis

Our analysis proceeded in two stages. First, we compared model classes under controlled conditions in which each task received the same number of observations. Second, we examined how different data collection strategies influenced model performance when the number and order of observations varied adaptively.

In the first stage, we evaluated DLVM against the IMLE approach. Observations were generated using the simulation procedure described in Figure 1, with each model receiving the same data under an equal allocation scheme (e.g., $n$ observations for each task). To evaluate parameter recovery (i.e., model accuracy), we computed the Kullback-Leibler Divergence (KLD) between estimated and ground-truth distributions. KLD was selected because it captures differences across the full distribution, rather than only summarizing lower-order moments. This analysis was designed to isolate the impact of modeling assumptions (i.e., nonlinear latent structure versus distribution independence) when the underlying data were the same.

In the second stage, we turned to the question of how sampling strategies that decide how to acquire observations of task performance affect parameter estimation. Unlike the first stage, here the models were not fit to identical data; instead, each model was applied to data generated under different collection procedures. We compared three strategies: Bayesian active learning with DALE, uniform random sampling and a fixed blocked test battery delivery. DALE selected trials sequentially by maximizing expected mutual information gain, continuing until 240 total observations were collected. The primer sequence consisted of 2 samples per task (a total of 16 samples). Random sampling drew trials uniformly across tasks without regard to informativeness. The standard Test Battery (TB) allocated a fixed number of observations per task, completing one task before moving to the next; to minimize ordering effects, task order was randomized across sessions.

For each sampling strategy, data were fit using either DLVM or IMLE. This design produced 6 configurations: DALE combined with DLVM (active sampling with a latent variable model), random sampling with DLVM, TB with DLVM, DALE combined with IMLE, random sampling with IMLE, and TB with IMLE. Comparing across these configurations allowed us to disentangle the effects of the modeling approach from those of the data collection strategy, and to assess how adaptivity in both modeling and sampling influences the efficiency and accuracy of parameter recovery. Below is the summary of the six configurations:

- **DLVM+DALE+PS2**: Data are collected adaptively using DALE with a primer sequence of 2 samples per task (a total of 16 observations). DALE is turned on after the 16th sample. The collected data are then fit with DLVM.
- **DLVM+RAND:** Data are collected by randomly selecting trials across tasks, without regard to informativeness. The collected data are then fit with DLVM.
- **DLVM+TB:** Data are collected using a fixed test battery, where tasks are completed sequentially with a fixed allocation of trials. The collected data are then fit with DLVM.
- **IMLE+DALE+PS2:** Data are collected using DALE with a primer sequence of 2 observations per task. The collected data are fitted with IMLE, which assumes independence across tasks.
- **IMLE+RAND:** Data are collected by randomly selecting trials across tasks. The resulting data are fit with IMLE.
- **IMLE+TB:** Data are collected using a fixed test battery. The resulting data are fit with IMLE.

A key assumption for this procedure is that ordering effects do not affect modeling results. When evaluating DLVM and DALE in human participants, this strong assumption was loosened by delivering a minimally sized mini-block of trials every time a new task was selected by the Bayesian algorithm, generally 4 identical or counterbalanced trials per task (Kasumba et al., 2025). That procedure resulted in high correspondence between the traditional block-style test battery delivery and the adaptive method. Appropriate minibatch size may be an additional hyperparameter worthy of exploration for optimal cross-task item delivery for human participants; it was essentially set to 1 for this study.

# Results

## Model Comparison under Fixed Observations

We first compared how well DLVM and IMLE fit the same data when each test received an equal number of observations. For example, allocating one observation per test meant that each task contributed a single draw from its ground-truth distribution. For tasks modeled with sigmoids, this observation was taken at a random length (e.g., a span length of 3 in the Corsi Simple Span task). **Figure 3** illustrates model fits for a representative session when two observations per test (16 total across eight tasks) as well as 50 observations per test (400 total) were available.

With only two observations per test, DLVM with latent dimensionality two (DLVM-2) consistently achieved lower KLD values (below 0.200) than IMLE across all tasks. This advantage was most pronounced for the span tasks, which require more independent data to estimate reliably due to their sigmoid structure. Both visually and quantitatively, DLVM provided more accurate estimates of task thresholds under these sparse data conditions. When 50 observations per test were available, IMLE shows superiority, recovering the ground-truth distributions almost perfectly. DLVM also performed well in this setting (KLD below 0.100), particularly in capturing the first moments while maintaining appropriate uncertainty in higher-order moments. This property illustrates a key feature of DLVM: by leveraging information across tasks, it avoids overconfidence while retaining flexible uncertainty representations. Overall DLVM maintains an advantage until after 7 observations per task (56 total), beyond which IMLE is better. Full results are reported in the Supplementary Material.

Validation performance demonstrates that DLVM models estimate marginal distributions more accurately than the IMLE method under these experimental conditions, particularly when only a limited number of observations per task are available. As shown in **Figure 4**, the DLVM-2 model requires roughly 20 observations per task (160 observations total) to achieve near-perfect accuracy (KLD ≈ 0) in estimating marginal distributions, whereas the IMLE method requires about 100 observations per task (800 total) to reach comparable accuracy. The advantage of DLVM over IMLE is most pronounced when fewer than 10 observations per task are available, highlighting its ability to make credible distributional parameter estimates under sparse data conditions by leveraging shared information across tasks and sessions. Furthermore, DLVM can estimate parameters for unobserved tests, a case where conventional approaches such as IMLE typically fail.

In this setting, the ground-truth sessions were generated from the latent space learned by DLVM. Provided that the search procedure is successful, DLVM is guaranteed in principle to recover the true latent position, consistent with our empirical findings. Other models (DLVM-1 and DLVM-3) exhibit comparable performance, with full results reported in the Supplementary Material.

## Effect of Sampling Strategy on Model Performance

**Figure 5** compares DLVM-2 and IMLE under different sampling strategies. For any given sampling strategy, DLVM outperforms IMLE, particularly when fewer than 60 observations are available. Within DLVM, DALE with a primer sequence achieves the lowest error, reducing KLD below 0.05 by ~80 observations and plateauing thereafter. DLVM with random sampling improves more gradually but reaches a very low error (KLD about 0.01) once all tasks are sufficiently represented. Table 2 shows a snapshot of the performance of each method after 20 observations, 80 observations and after 240 observations.

The advantage of DALE stems from its adaptive allocation of trials. As shown in **Figure 6**, DALE tends to concentrate sampling on complex distributional tasks while allocating fewer trials to accuracy tasks, implying that the former are generally more informative. This adaptive behavior varies across sessions, with each of the nine tested requiring a distinct sampling profile for their responses on a given simulated test session, highlighting DALE's ability to personalize test administration.

By contrast, IMLE+TB exhibited the worst accuracy. Because IMLE cannot estimate parameters for unobserved tests, its accuracy improved only in stepwise increments as tasks were added sequentially, remaining far higher in KLD than DLVM at intermediate points (e.g., at 80 observations, DLVM+TB = 0.148 vs. IMLE+TB = 39.8). IMLE with random sampling performed somewhat better but consistently lagged behind DLVM. To assess the extent to which these findings may have depended on our selection of a DLVM-based generative process, we repeated the analysis using an IMLE-based generative process. Results provided in the Supplementary Material showed the same qualitative pattern, with DLVM/DALE retaining an advantage in the sparse-data regime, whereas IMLE performed best when sufficient data were available under its own generative specification. Overall, these results indicate that DLVM substantially improves estimation under limited data and that its performance is further enhanced by adaptive sampling strategies such as DALE. This pattern was consistent across DLVMs with different latent dimensionalities.

## Latent Space Dynamics

Examining DALE's trajectories in the latent space shows that it tends to make large adjustments during the initial trials, followed by convergence to a localized region of latent space after approximately 30 observations (**Figure 7**). Beyond this point, updates are minor and tend to remain within the same region. Consistently, DALE converges to regions of high log probability, specifically, areas with normalized negative log probability values below about 0.04.

Negative log probability is computed by generating 240 simulated data points per task from the ground-truth latent position and evaluating their likelihood under the candidate position. Because the latent space is nonlinear and non-convex, the final position identified by DALE does not always coincide with the true latent position. For example, session LD2-086 (bottom left panel of Figure 7) ends with a Root Mean Squared Error (RMSE) of 5.46 relative to the ground truth, yet achieves a normalized negative log probability of 0.001. This result occurs because multiple regions of the latent space yield equally high-probability reconstructions of the observed data, and the search procedure ended in one of these away from the true latent position.

In other words, the nonlinear structure of the latent space admits several plausible solutions that can generate similar observable profiles. Figure 7 visualizes these alternative regions for the representative simulated sessions, while **Figure 8** quantifies the effect: only seven out of 88 sessions converged to positions with a negative log probability above 0.05 (mean = 0.0133). Overall, the mean RMSE across all sessions was 1.02. These findings underscore that DALE's convergence reflects probability structure rather than strict positional accuracy in the latent space, at least for this configuration of DLVM.

# Discussion

Our findings demonstrate that nonlinear Distributional Latent Variable Models (DLVMs), particularly when paired with adaptive sampling through Distributional Active LEarning (DALE), offer clear advantages for cognitive testing in data-limited settings. By leveraging cross-task and cross-individual information, DLVMs produced credible task output distributional parameter estimates (e.g., mean response times, accuracies, psychometric thresholds) with only a handful of observations while maintaining appropriate uncertainty. In contrast, IMLE required substantially more data to stabilize and could not accommodate missing task data, highlighting the robustness of DLVM in sparse and incomplete data regimes.

These results point to an important trade-off. DLVMs excel when data are scarce, efficiently exploiting shared structure to generate reliable estimates. IMLE benefits from larger sample sizes and would generally be expected to surpass DLVM once sufficient observations are available (perhaps a very large number) because it is the more flexible model. This scenario is the opposite of the most common use of machine learning algorithms, where highly flexible models such as neural networks need large amounts of data to converge to accurate representations of underlying data-generating processes. By building a somewhat less flexible machine learning model able to compress multiple variable interrelationships into a low-order nonlinear embedding, we are able to perform meaningful inference with fewer data.

Our comparison of different sampling strategies reinforces this theme. DALE accelerated convergence toward accurate estimates by targeting the most informative trials, providing clear benefits in the early stages of testing. Random sampling eventually surpassed active sampling performance, but only after substantially more observations, underscoring the inefficiency of uninformed data collection. This is an interesting observation, however, because it suggests that for at least some machine learning EF models, a transition to random sampling once active learning plateaus may reveal additional structure in the data. When sample counts grow large, DALE does tend to start oversampling a subset of tasks, so refining the acquisition function may have the same effect. Fixed test batteries performed worst with IMLE, as expected, but DLVM retained the ability to infer across missing tasks, softening the limitations of the standard rigid test designs. This is evident with the TB sampling strategy at 80 total observations (Table 2), where only three tasks have been sampled: DLVM+TB achieved a lower KLD (0.148) than IMLE+TB (39.8). Because DLVM can borrow information across tasks, it can still produce more accurate estimates for tasks with missing observations, demonstrating greater robustness to incomplete data than IMLE. These results illuminate the importance of adaptive and individualized data collection strategies, especially when testing time is constrained.

Finally, our analysis of DALE's latent-space search shows that the algorithm made large adjustments early, then typically refined estimates with smaller updates after about 30 trials. It consistently converged to regions of high probability even when these did not coincide exactly with the ground-truth latent position. This result reflects the nonlinearity of the latent space itself rather than a limitation of DALE, where multiple regions can yield equivalent reconstructions of observed behavior. Importantly, DALE uncovered these high-probability regions with relatively few trials, reinforcing its functional utility for individualized assessment and adaptive testing, while not necessarily guaranteeing recovery of a unique underlying latent psychological trait profile for a given individual. In situations where positional accuracy is vital, the latent spaces need to be regularized better to ensure convexity or near-convexity. Ongoing work in machine learning research focuses on such regularization (Lee & Park, 2023). We used a neural network-based latent space, but machine learning models such as Gaussian processes (GP) would be expected generate smoother latent spaces (Marticorena, Wong, Browning, Wilbur, Davey, et al., 2024; Marticorena, Wong, Browning, Wilbur, Jayakumar, et al., 2024). Important to note that these approaches may improve the identifiability of the latent space, they do not eliminate the challenge entirely.

We acknowledge that using the DLVM-learned latent space as the primary generative process introduces a structural advantage for DLVM over IMLE. This design was chosen because it provides a controlled setting in which both latent-space recovery and recovery of observable distributional parameters can be evaluated under cross-task dependence. To examine the impact of this choice, we repeated the analysis using IMLE fits to the real human item-level data as the generative process, with each individual represented by their IMLE-fitted parameter values; these results are provided in the Supplementary Material. Under this alternative setup, IMLE unsurprisingly performs best when recovering parameters from data generated by the IMLE model, particularly once sufficient observations are available. Even so, DLVM and DALE continue to show an advantage in the sparse-data regime, and the overall pattern remains qualitatively consistent with the primary results.

Regarding the original hypotheses, we can say they were both supported by the study results, with DLVM alone and DLVM+DALE providing more inference per observation than the tested alternatives. The number of observations for the more flexible IMLE estimator to match the 2D DLVM estimator was higher than expected, at over 800 for these testing conditions. These performance disparities may differ under other testing conditions and model configurations, but the potential for the newer methods to efficiently provide accurate executive function estimates for individuals is confirmed by the study results.

An important consideration of the present study is that while it, as with most machine learning studies, emphasized prediction over explanation, the nature of DLVM and its data-driven dimensionality reduction provides an avenue to explanation, as well. The outliers in latent variable space that obtained good distributional predictions (i.e., the sessions plotted on the righthand side of Figure 8) occurred because the learning optimizer found a local minimum in the nonconvex space it was computing over. If latent space representation and explanatory power were important for a particular application, a straightforward option would be to swap out a kernel method for the neural network and force a convex space with monotonic mappings, as described above. Other options include regularizing more heavily to enforce monotonic latent representations and training on large population data to achieve the same effect. The elegance of this method is that all these options are available and are completely modular in the sense that DALE can operate in conjunction with any of them. Therefore, this procedure is a generalization of existing approaches that offers considerable utility for experimental paradigms, individual differences research, individual phenotyping, etc.

DALE could support adaptive cognitive assessment in settings where testing time is limited or where administering a fixed battery would place unnecessary burden on examinees. Potential applications include clinical screening, longitudinal monitoring, and large-scale studies requiring reliable estimates of performance across several executive-function tasks. As illustrated in Figure 9, an examinee first completes a short primer spanning the task battery. DALE the updates it's belief over the internal latent representation. Using this posterior belief, it selects the task or item expected to provide the greatest reduction in uncertainty, updates its estimates after each response, and repeats this process until a prespecified precision or time criterion is reached. One examinee might therefore receive additional Stroop trials, whereas another might receive more span or countermanding trials, depending on where their performance remains most uncertain. The final output is an uncertainty-aware summary of observable performance, such as accuracy, response-time distributions, or psychometric thresholds. This adaptive functionality remains useful even when the DLVM latent representation is not uniquely identifiable, because distinct latent solutions may support similar predictions and lead to the same testing decisions.

One important consideration for future development is that DLVM was explicitly designed to incorporate data as currently collected for common cognitive test batteries. This means that the behavioral data are often highly reductionist and decontextualized with many repeated trials. Much more natural acquisition functions can be designed for true feature spaces, where any combination of underlying features can be instantiated into a task item (Song et al., 2018). We have found that optimal sampling in such scenarios virtually never repeats task items, most likely because a previously unsampled region of feature space is likely to be much more informative than a previously sampled region. Therefore, we believe that future behavioral tasks, when paired with advanced machine learning or artificial intelligence algorithms, should be designed to be multidimensional, contextualized and full-featured in order to fully exploit these methodological advances. We describe one step in this direction with a companion article in this issue.

In sum, DLVMs provide a flexible and efficient framework for modeling cognition under sparse and incomplete data, and DALE enhances their utility by ensuring that each additional trial maximizes information. Together, they support a more adaptive, individualized approach to cognitive testing that is both resource-efficient and theoretically grounded.

# Conclusions

This study builds on our prior work by evaluating how distributional latent variable models (DLVM) and Bayesian Distributional Active Learning (DALE) perform relative to conventional maximum likelihood estimation (IMLE) using simulations with known ground truth. We found that DLVM provides a clear advantage under sparse data conditions, producing credible parameter estimates even when only a handful of observations are available. This benefit was most pronounced for span tasks, which are modeled with sigmoidal functions and typically require more data to estimate reliably. While IMLE distribution-parameter estimation accuracy eventually surpassed that of DLVM with larger data sets, DLVM retained the critical advantage of inferring parameters for unobserved tasks by incorporating informative trends across tests and individuals.

Our second set of analyses demonstrated that DALE further enhances testing efficiency by guiding trial selection adaptively. By prioritizing the most informative observations, DALE accelerated convergence to accurate estimates, particularly in the first 30 trials. Random sampling and fixed test batteries required substantially more data to achieve comparable accuracy, with IMLE under fixed batteries showing the weakest performance due to its inability to handle unobserved tests. DLVM, in contrast, maintained robust inference even under incomplete data, highlighting the synergy between DLVM and DALE.

Taken together, these findings suggest a promising framework for shorter, adaptive and individualized cognitive assessments. DLVM allows trial-level information sharing across tasks and participants, while DALE provides a principled sampling strategy that reduces data requirements without sacrificing precision.

# Declarations

## Funding

The research reported here was supported by T32NS115672 and the EF+Math Program of the Advanced Education Research and Development Fund (AERDF) through funds provided to Washington University in St. Louis. The opinions expressed are those of the authors and do not necessarily represent the views of the EF+Math Program or AERDF.

## Conflicts of Interest/Competing Interests

The authors declare that they have no conflicts of interest or competing interests.

## Ethics Approval

All procedures performed in this study were approved by the Institutional Review Board at the University of California, Riverside, and adhered to the ethical standards of the 1964 Declaration of Helsinki and its later amendments or comparable ethical standards.

## Consent to Participate

Written informed consent was obtained from all participants prior to their inclusion in the study.

## Consent for Publication

Not applicable.

## Availability of Data and Materials

Project information, including the data necessary to replicate these analyses and supplementary material, can be found at https://osf.io/ynbdr/.

## Code Availability

The code necessary to replicate these analyses can be referenced at https://osf.io/ynbdr/.

## Authors' Contributions

Robert Kasumba: Conception and design of the study, data analysis, manuscript drafting.

Dom CP Marticorena, Zeyu Lu: Data analysis, manuscript editing.

Mingyang Zhong, Paul Beggs: Data analysis.

Anja Pahor: Oversight of original human data collection.

Geetha Ramani, Imani Goffney: equitable experimental design.

Susanne M Jaeggi, Aaron R Seitz, Jacob R Gardner: Overall supervision, secured funding, manuscript editing.

Dennis L. Barbour: Overall supervision, secured funding, manuscript finalization.

All authors read and approved the final manuscript.

## Open Practices

The authors followed guidelines recommended by the American Psychological Association (APA) for reporting methods and analyses.

# Figures and Tables

**Table 1**

*Cognitive tasks modeled, the specific EFs they are designed to measure, as well as the distributions of their data-generating processes. The main parameter of interest for each task is presented in bold. The secondary parameter is also provided to capture the full distributions.*

| Task | Executive Functions Reflected | Item-level Units | Modeled Distribution | Parameters of Interest |
|---|---|---|---|---|
| Corsi Complex Span | working memory, inhibitory control | binary accuracy at each length | psychometric sigmoid | **threshold**, spread |
| Corsi Simple Span | working memory, inhibitory control | binary accuracy at each length | psychometric sigmoid | **threshold**, spread |
| Countermanding | inhibitory control | response time | lognormal | **mean**, standard deviation |
| Numerical Stroop | inhibitory control | response time | lognormal | **mean**, standard deviation |
| Running span (Length 2) | working memory | binary accuracy | binomial | **probability** |
| Running span (Length 3) | working memory | binary accuracy | binomial | **probability** |
| PASAT+ | sustained attention, cognitive flexibility | binary accuracy | binomial | **probability** |
| Cancellation | selective attention, inhibitory control | binary accuracy | binomial | **probability** |

**Table 2**

*Performance (KLD) of Each Method after the Given Number of Observations*

| **Method** | **20 observations** | **80 observations** | **240 observations** |
|---|---|---|---|
| DLVM-2 + TB | 0.318 ± 0.279 | 0.148 ± 0.206 | 0.0146 ± 0.0469 |
| DLVM-2 + Random | 0.119 ± 0.124 | **0.0304 ± 0.0444** | **0.00600 ± 0.00900** |
| DLVM-2 + DALE | **0.0874 ± 0.105** | **0.0304 ± 0.0592** | 0.0191 ± 0.0363 |
| IMLE + TB | 68.1 ± 27.5 | 39.8 ± 29.5 | 0.0291 ± 0.0134 |
| IMLE + Random | 4.40 ± 11.6 | 0.101 ± 0.0630 | 0.0313 ± 0.0139 |
| IMLE + DALE | 0.332 ± 0.180 | 0.247 ± 0.169 | 0.227 ± 0.168 |

*Note*. Values represent mean KLD ± standard deviation. Lower is better. Bold means the best method after the corresponding number of observations.

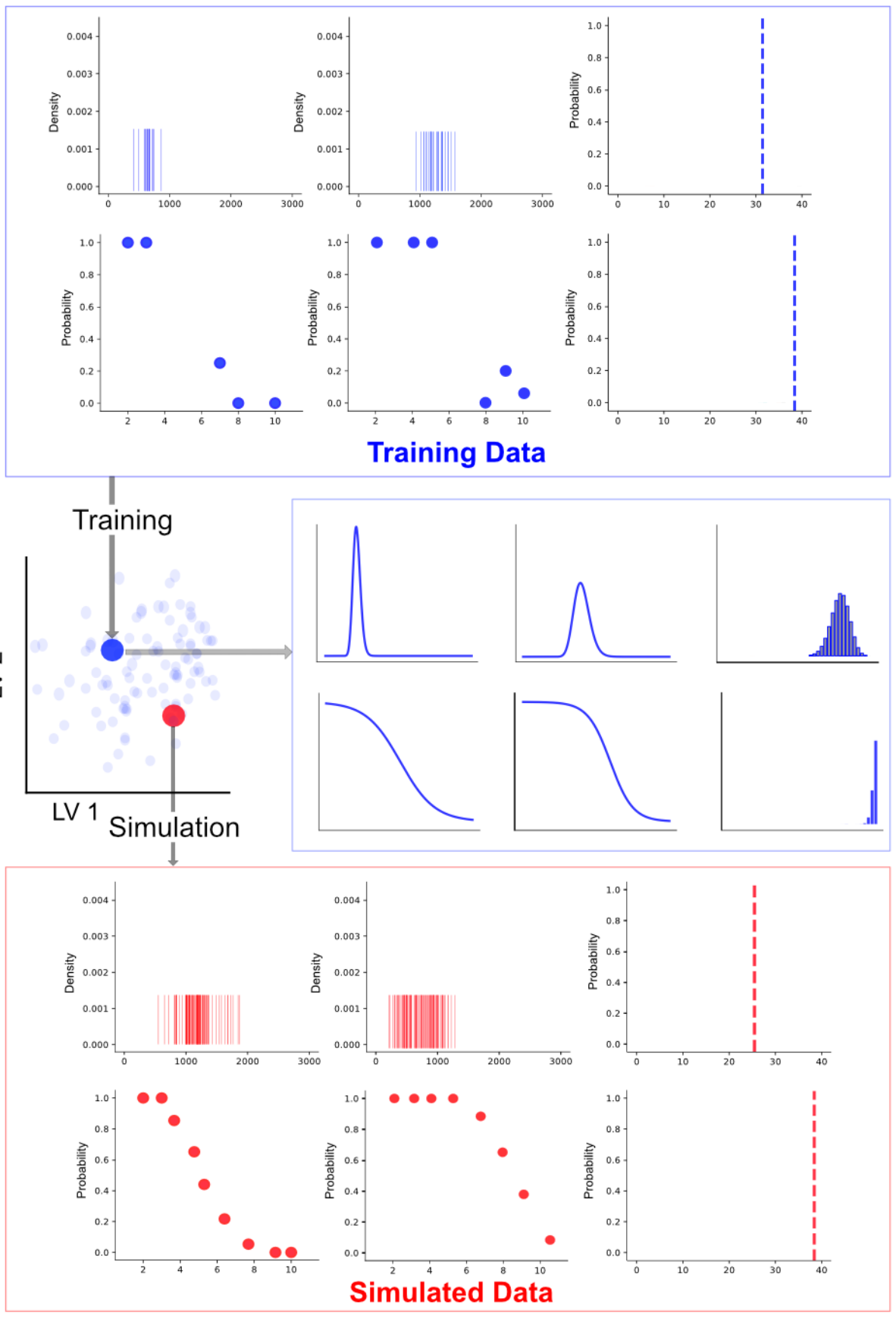

**Figure 1**: Procedure for generating a simulated dataset. DLVM-2 model was trained using the COLL10 dataset (*n* = 88 individual sessions) to learn the latent space. This space was then grid-sampled to create simulated sessions (*n* = 88). The simulated dataset was used for all further analysis. LV 1 and LV 2 refer to the Latent Variables in the first and second dimension, respectively. Distribution units are native to the underlying task.

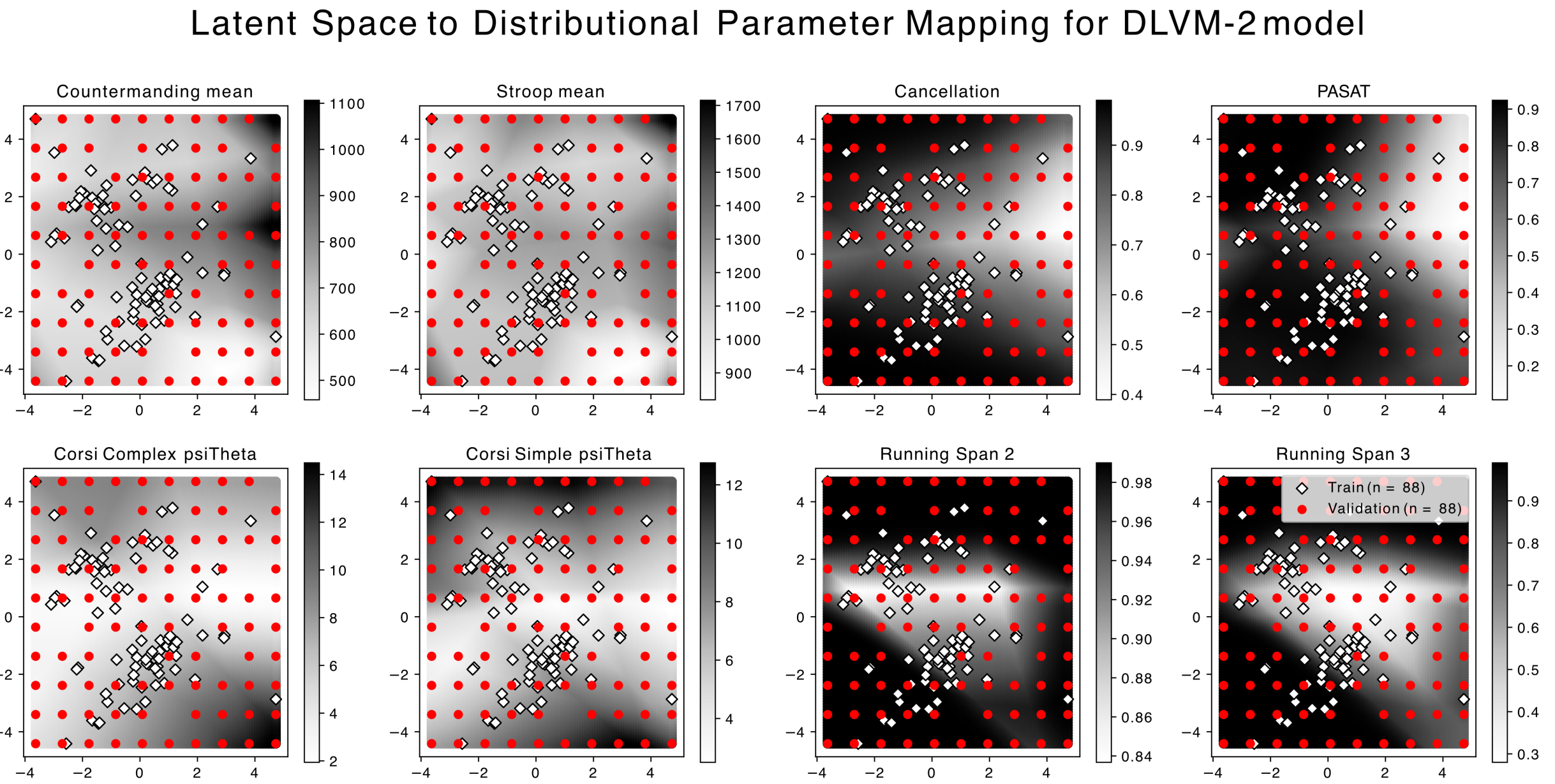

**Figure 2**: Latent spaces showing ground truth latent variables (plot symbols) and mappings to marginals (grayscales). The abscissas and ordinates represent the latent dimensions of the learned DLVM-2 model. Correlations are apparent between different tasks. For instance, sessions with slower response times on Stroop and countermanding tend to have lower working memory thresholds (Corsi Simple and Complex $\psi_\theta$). Latent space units are arbitrary; grayscale distribution-mapping units are native to the underlying task.

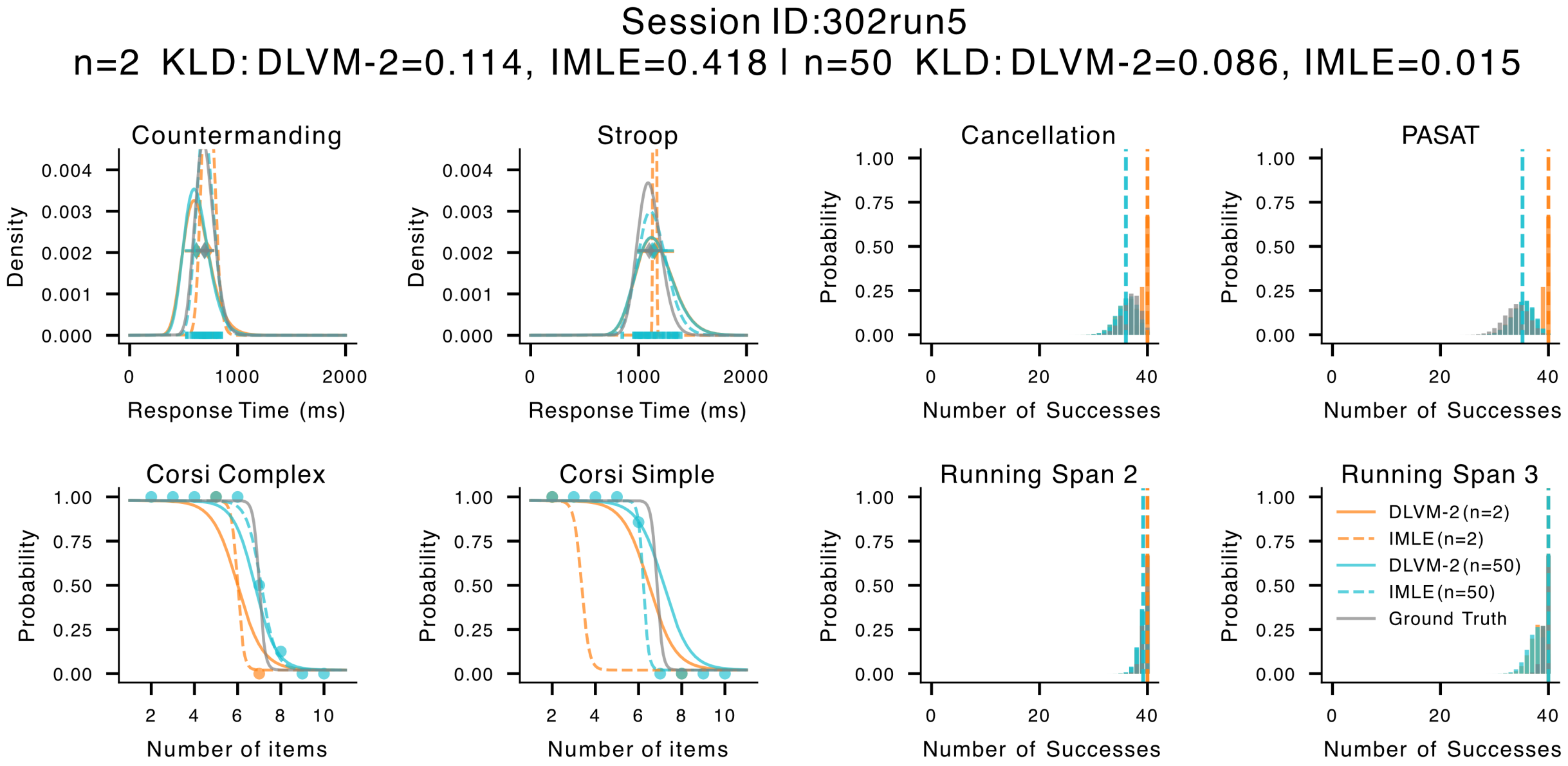

**Figure 3**: Marginal fits on the training data for a median-fit session after 2 observations and after 50 observations per task. DLVM fits are indicated by solid lines, and IMLE fits by dashed lines. The ground truth generative model is indicated by gray lines. For visualization purposes of binary accuracy tasks, we assume $n = 40$ repeats for the binomial distribution.

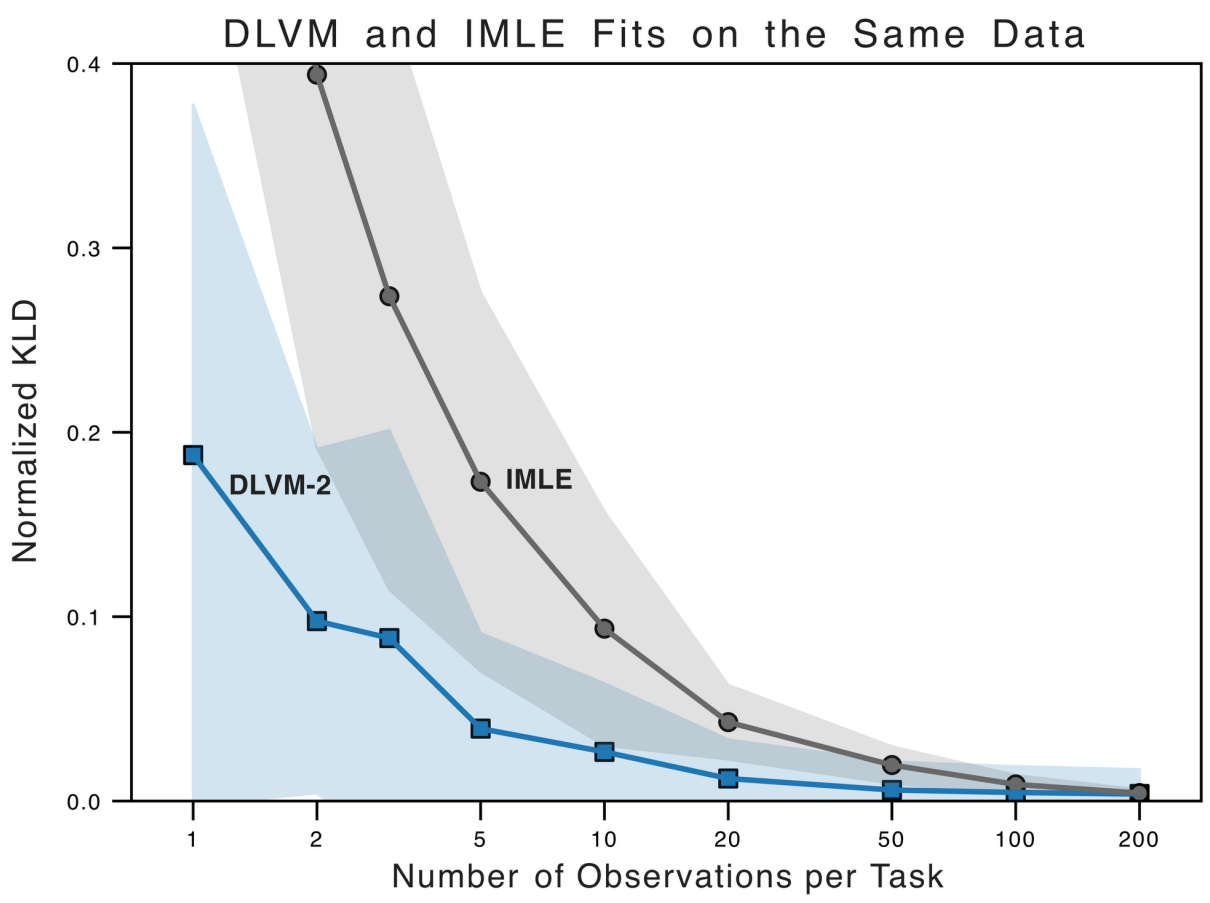

**Figure 4**: Mean ± standard deviation of DLVM and IMLE model accuracy ($n$ = 88 each) as the number of observations per task increases.

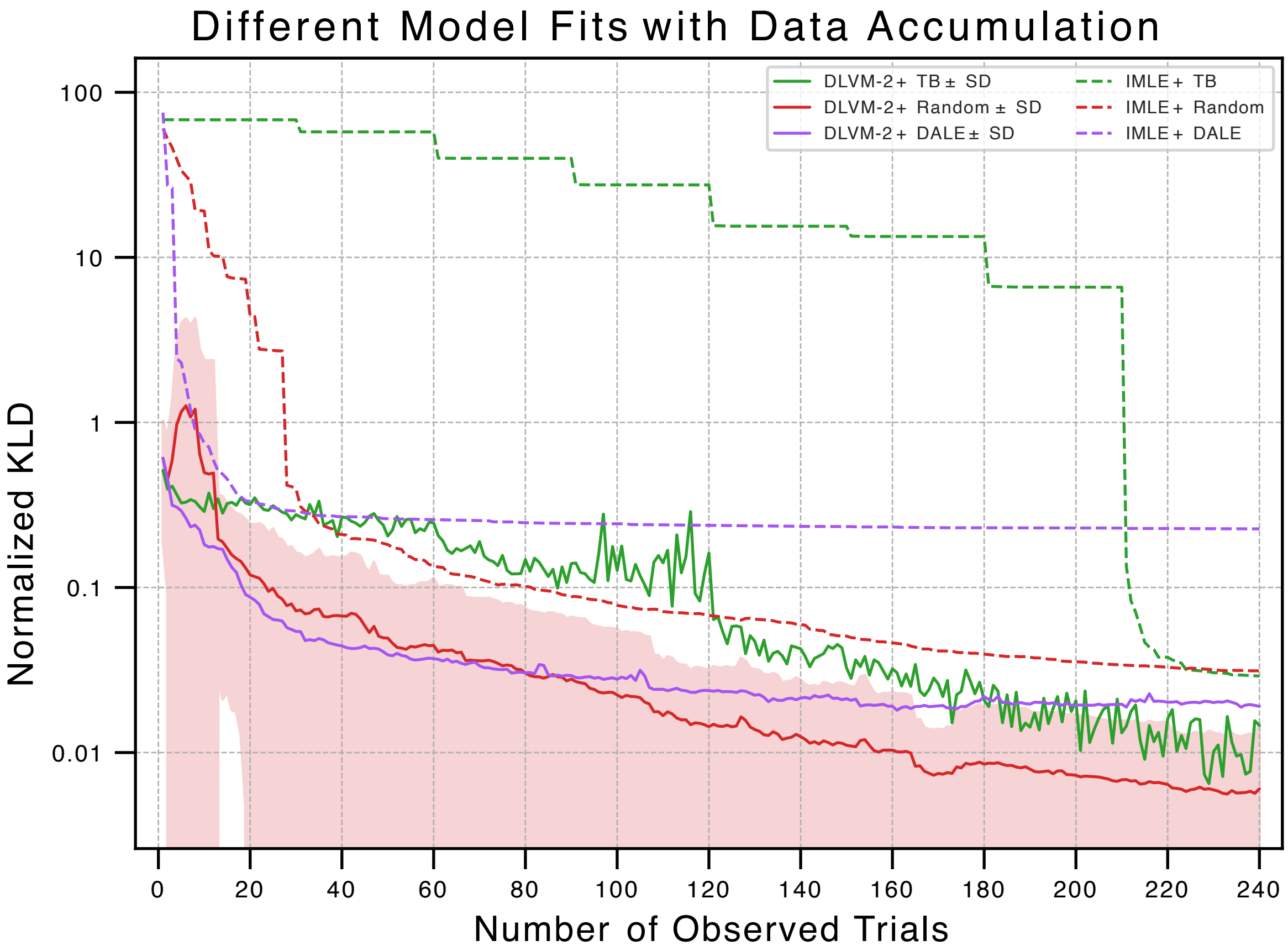

**Figure 5**: Performance of the different estimation methods and the sampling strategies.

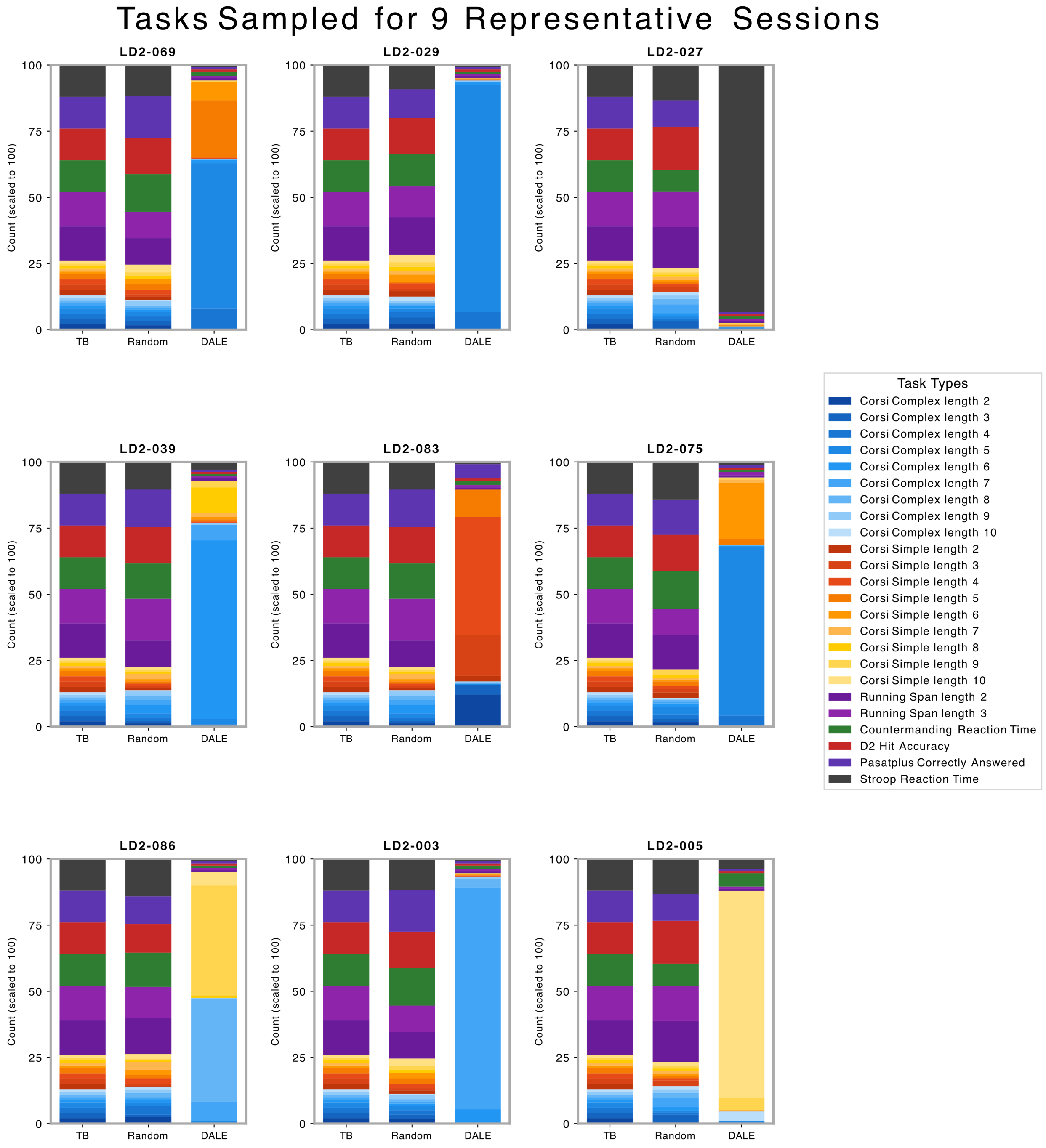

**Figure 6**: Distributions of the first 100 test items delivered by the DLVM-2+DALE by cognitive test for a few representative sessions in the validation set. Each session consists of a unique battery in a unique order, depending on which tasks are most informative for estimating the cognitive performance in that session.

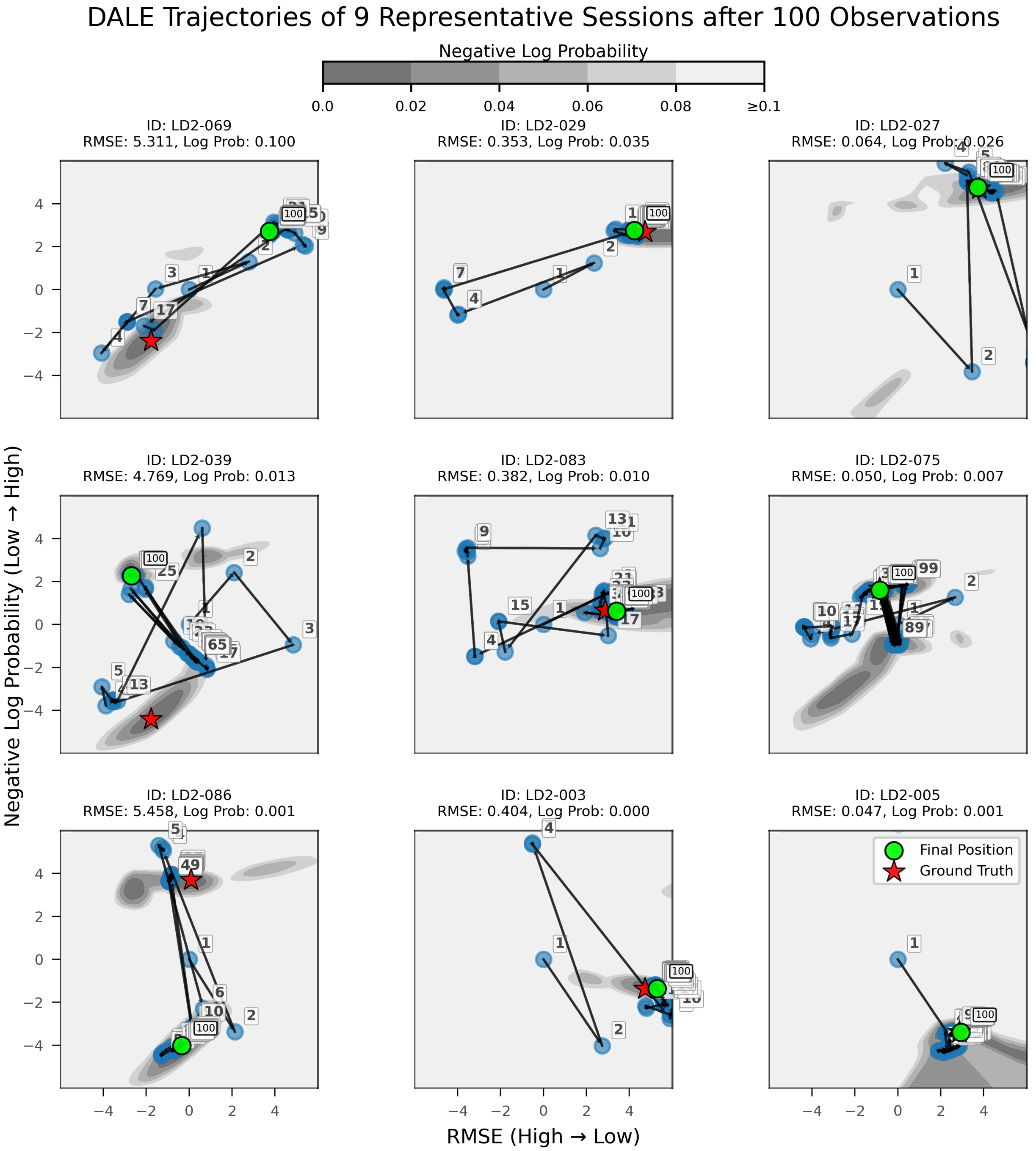

**Figure 7**: DLVM-2+DALE latent position updates as more data is collected for nine representative simulated sessions based on the RMSE values between the ground truth position and the position estimated by DALE after accumulating 100 observations.

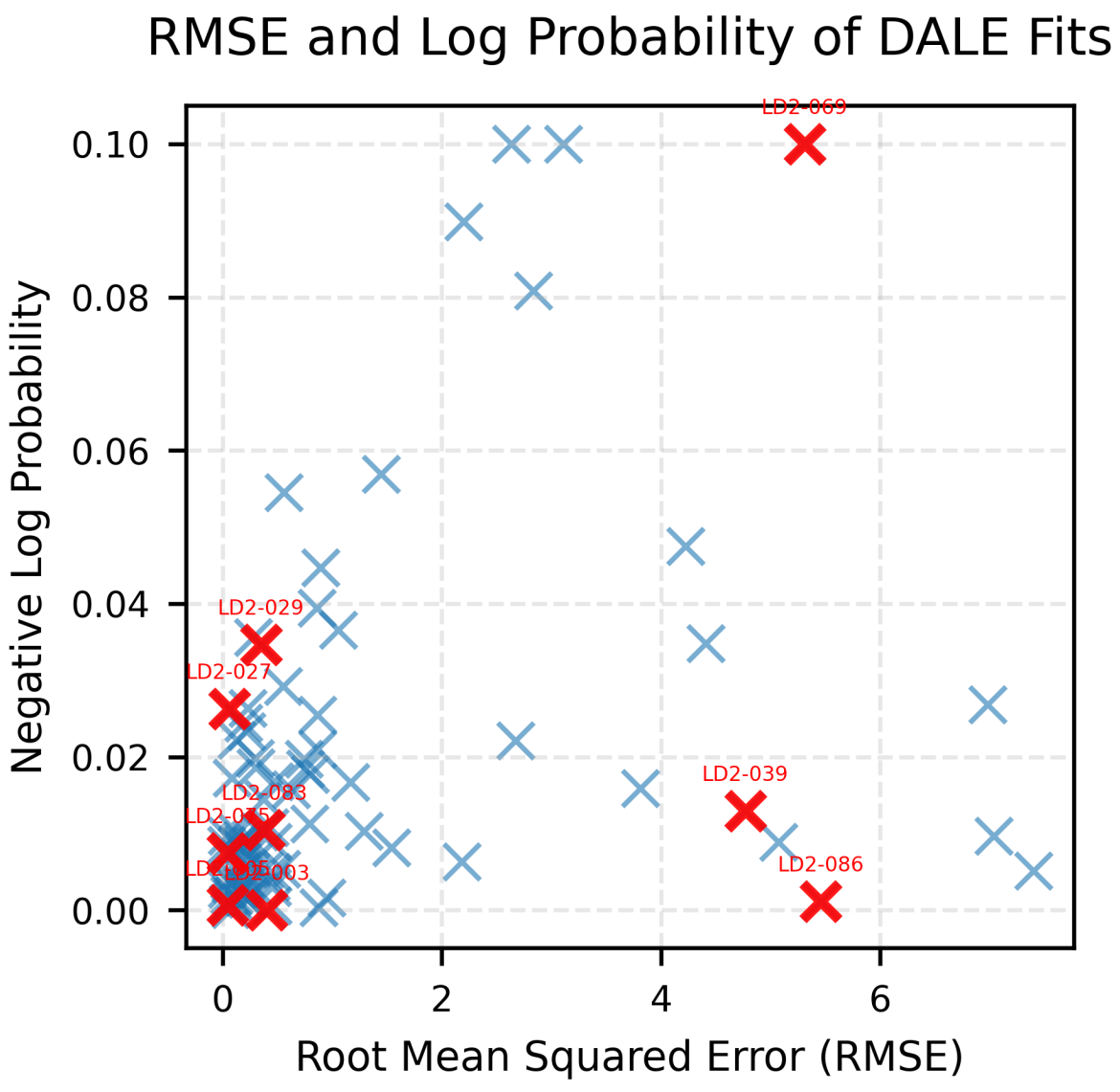

**Figure 8**: Distribution of latent position RMSE values, latent positions and the ground truth latent positions and normalized negative log probability values for the final DLVM-2+DALE estimates. Red symbols indicate the examples from Figures 6 and 7; blue symbols, all remaining validation sessions.

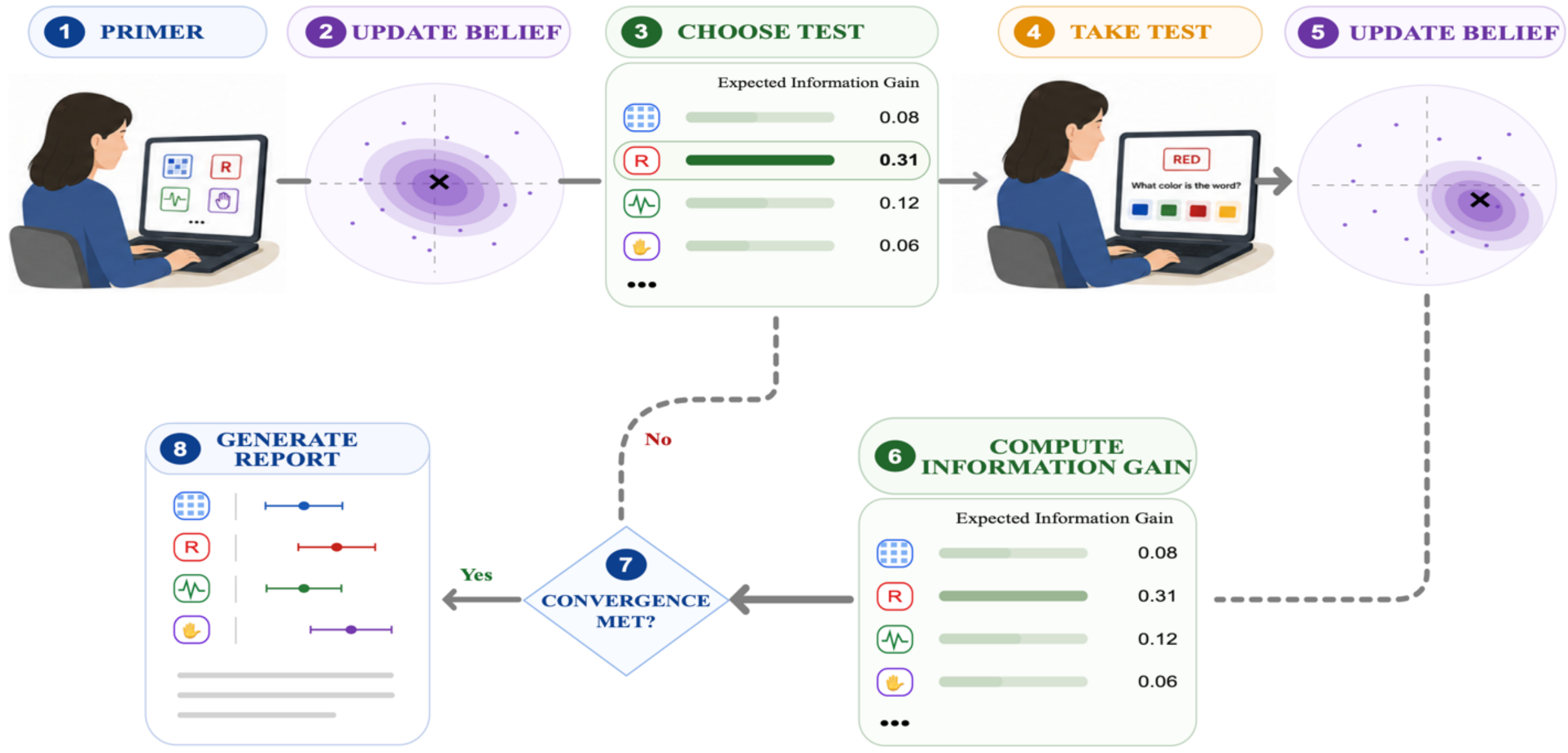

**Figure 9**: Illustration of DALE in an applied adaptive assessment. The process starts with a short primer sequence spanning the available tasks. DALE updates the posterior belief over latent representations and selects the next task item expected to provide the greatest information gain. Responses are incorporated sequentially until a fixed trial budget or uncertainty-based stopping criterion is reached.

References

Barbour, D. L., DiLorenzo, J. C., Sukesan, K. A., Song, X. D., Chen, J. Y., Degen, E. A., Heisey, K. L., & Garnett, R. (2019). Conjoint psychometric field estimation for bilateral audiometry. *Behavior Research Methods*, *51*(3), 1271–1285. https://doi.org/10.3758/s13428-018-1062-3

Brunetti, R., Del Gatto, C., & Delogu, F. (2014). eCorsi: Implementation and testing of the Corsi block-tapping task for digital tablets. *Frontiers in Psychology*, *5*, 939. https://doi.org/10.3389/fpsyg.2014.00939

Chesley, B., & Barbour, D. L. (2020). Visual field estimation by probabilistic classification. *IEEE Journal of Biomedical and Health Informatics*, *24*(12), 3499–3506. https://doi.org/10.1109/JBHI.2020.2999567

Cousineau, D., Brown, S., & Heathcote, A. (2004). Fitting distributions using maximum likelihood: Methods and packages. *Behavior Research Methods, Instruments, & Computers*, *36*(4), 742–756. https://doi.org/10.3758/BF03206555

Cox, M., & de Vries, B. (2021). Bayesian pure-tone audiometry through active learning under informed priors. *Frontiers in Digital Health*, *3*, 723348. https://doi.org/10.3389/fdgth.2021.723348

Cragg, L., & Gilmore, C. (2014). Skills underlying mathematics: The role of executive function in the development of mathematics proficiency. *Trends in Neuroscience and Education*, *3*(2), 63–68. https://doi.org/10.1016/j.tine.2013.12.001

Diamond, A. (2013). Executive functions. *Annual Review of Psychology*, *64*, 135–168. https://doi.org/10.1146/annurev-psych-113011-143750

Embretson, S. E. (1991). A multidimensional latent trait model for measuring learning and change. *Psychometrika*, *56*(3), 495–515. https://doi.org/10.1007/BF02294487

Friedman, N. P., & Miyake, A. (2017). Unity and diversity of executive functions: Individual differences as a window on cognitive structure. *Cortex*, *86*, 186–204. https://doi.org/10.1016/j.cortex.2016.04.023

Gutiérrez-Martínez, F., Ramos-Ortega, M., & Vila-Chaves, J.-Ó. (2018). Eficacia ejecutiva en tareas de interferencia tipo Stroop. Estudio de validación de una versión numérica y manual (CANUM). *Anales de Psicología*, *34*(1), 184. https://doi.org/10.6018/analesps.34.1.263431

Hedge, C., Powell, G., & Sumner, P. (2018). The reliability paradox: Why robust cognitive tasks do not produce reliable individual differences. *Behavior Research Methods*, *50*(3), 1166–1186. https://doi.org/10.3758/s13428-017-0935-1

Heisey, K. L., Buchbinder, J. M., & Barbour, D. L. (2018). Concurrent bilateral audiometric inference. *Acta Acustica United with Acustica*, *104*(5), 762–765. https://doi.org/10.3813/AAA.919218

Heisey, K. L., Walker, A. M., Xie, K., Abrams, J. M., & Barbour, D. L. (2020). Dynamically masked audiograms with machine learning audiometry. *Ear and Hearing*, *41*(6), 1692–1702. https://doi.org/10.1097/AUD.0000000000000891

Kasumba, R., Marticorena, D. C. P., Pahor, A., Ramani, G., Goffney, I., Jaeggi, S. M., Seitz, A. R., Gardner, J. R., & Barbour, D. L. (2025). Distributional latent variable models with an application in active cognitive testing. *IEEE Transactions on Cognitive and Developmental Systems*, 1–11. https://doi.org/10.1109/TCDS.2025.3548962

Kessels, R. P. C., van den Berg, E., Ruis, C., & Brands, A. M. A. (2008). The backward span of the Corsi Block-Tapping Task and its association with the WAIS-III Digit Span. *Assessment*, *15*(4), 426–434. https://doi.org/10.1177/1073191108315611

Kingma, D. P., & Ba, J. (2017). *Adam: A method for stochastic optimization* (arXiv:1412.6980). arXiv. https://doi.org/10.48550/arXiv.1412.6980

Lee, Y., & Park, F. C. (2023). On explicit curvature regularization in deep generative models. *Proceedings of 2nd Annual Workshop on Topology, Algebra, and Geometry in Machine Learning (TAG-ML)*, *221*, 505–518. https://proceedings.mlr.press/v221/lee23a.html

Linkovski, O., Rodriguez, C. I., Wheaton, M. G., Henik, A., & Anholt, G. E. (2021). Momentary induction of inhibitory control and its effects on uncertainty. *Journal of Cognition*, *4*(1), 10. https://doi.org/10.5334/joc.133

Löffler, C., Frischkorn, G. T., Hagemann, D., Sadus, K., & Schubert, A.-L. (2024). The common factor of executive functions measures nothing but speed of information uptake. *Psychological Research*, *88*(4), 1092–1114. https://doi.org/10.1007/s00426-023-01924-7

Marticorena, D. C. P., Wong, Q. W., Browning, J., Wilbur, K., Davey, P. G., Seitz, A. R., Gardner, J. R., & Barbour, D. L. (2024). Active mutual conjoint estimation of multiple contrast sensitivity functions. *Journal of Vision*, *24*(8), 6. https://doi.org/10.1167/jov.24.8.6

Marticorena, D. C. P., Wong, Q. W., Browning, J., Wilbur, K., Jayakumar, S., Davey, P. G., Seitz, A. R., Gardner, J. R., & Barbour, D. L. (2024). Contrast response function estimation with nonparametric Bayesian active learning. *Journal of Vision*, *24*(1), 6. https://doi.org/10.1167/jov.24.1.6

Miyake, A., Friedman, N. P., Emerson, M. J., Witzki, A. H., Howerter, A., & Wager, T. D. (2000). The unity and diversity of executive functions and their contributions to complex "Frontal Lobe" tasks: A latent variable analysis. *Cognitive Psychology*, *41*(1), 49–100. https://doi.org/10.1006/cogp.1999.0734

Pahor, A., Collins, C., Smith-Peirce, R. N., Moon, A., Stavropoulos, T., Silva, I., Peng, E., Jaeggi, S. M., & Seitz, A. R. (2021). Multisensory facilitation of working memory training. *Journal of Cognitive Enhancement*, *5*, 386–395. https://doi.org/10.1007/s41465-020-00196-y

Pahor, A., Mester, R. E., Carrillo, A. A., Ghil, E., Reimer, J. F., Jaeggi, S. M., & Seitz, A. R. (2022). UCancellation: A new mobile measure of selective attention and concentration. *Behavior Research Methods*, *54*, 2602–2617. https://doi.org/10.3758/s13428-021-01765-5

Pahor, A., Stavropoulos, T., Jaeggi, S. M., & Seitz, A. R. (2019). Validation of a matrix reasoning task for mobile devices. *Behavior Research Methods*, *51*(5), 2256–2267. https://doi.org/10.3758/s13428-018-1152-2

Ratcliff, R., & McKoon, G. (2008). The diffusion decision model: Theory and data for two-choice decision tasks. *Neural Computation*, *20*(4), 873–922. https://doi.org/10.1162/neco.2008.12-06-420

Ratcliff, R., & Tuerlinckx, F. (2002). Estimating parameters of the diffusion model: Approaches to dealing with contaminant reaction times and parameter variability. *Psychonomic Bulletin & Review*, *9*(3), 438–481. https://doi.org/10.3758/BF03196302

Rojo, M., Maddula, P., Fu, D., Guo, M., Zheng, E., Grande, Á., Pahor, A., Jaeggi, S., Seitz, A., Goffney, I., Ramani, G., Gardner, J. R., & Barbour, D. (2023). *Scalable probabilistic modeling of working memory performance*. PsyArXiv. https://doi.org/10.31234/osf.io/nq6yg

Rouder, J. N., & Haaf, J. M. (2019). A psychometrics of individual differences in experimental tasks. *Psychonomic Bulletin & Review*, *26*(2), 452–467. https://doi.org/10.3758/s13423-018-1558-y

Rouder, J. N., Lu, J., Speckman, P., Sun, D., & Jiang, Y. (2005). A hierarchical model for estimating response time distributions. *Psychonomic Bulletin & Review*, *12*(2), 195–223. https://doi.org/10.3758/BF03257252

Scarpina, F., & Tagini, S. (2017). The Stroop Color and Word Test. *Frontiers in Psychology*, *8*, 557. https://doi.org/10.3389/fpsyg.2017.00557

Schlittenlacher, J., Turner, R. E., & Moore, B. C. J. (2018). Audiogram estimation using Bayesian active learning. *The Journal of the Acoustical Society of America*, *144*(1), 421–430. https://doi.org/10.1121/1.5047436

Song, X. D., Garnett, R., & Barbour, D. L. (2017). Psychometric function estimation by probabilistic classification. *The Journal of the Acoustical Society of America*, *141*(4), 2513. https://doi.org/10.1121/1.4979594

Song, X. D., Sukesan, K. A., & Barbour, D. L. (2018). Bayesian active probabilistic classification for psychometric field estimation. *Attention, Perception, & Psychophysics*, *80*(3), 798–812. https://doi.org/10.3758/s13414-017-1460-0

Song, X. D., Wallace, B. M., Gardner, J. R., Ledbetter, N. M., Weinberger, K. Q., & Barbour, D. L. (2015). Fast, continuous audiogram estimation using machine learning. *Ear and Hearing*, *36*(6), e326–e335. https://doi.org/10.1097/AUD.0000000000000186

Twinomurinzi, H., Myburgh, H., & Barbour, D. L. (2024). Active transfer learning for audiogram estimation. *Frontiers in Digital Health*, *6*, 1267799. https://doi.org/10.3389/fdgth.2024.1267799

Veenman, M., Stefan, A. M., & Haaf, J. M. (2024). Bayesian hierarchical modeling: An introduction and reassessment. *Behavior Research Methods*, *56*(5), 4600–4631. https://doi.org/10.3758/s13428-023-02204-3

Younger, J. W., O'Laughlin, K. D., Anguera, J. A., Bunge, S. A., Ferrer, E. E., Hoeft, F., McCandliss, B. D., Mishra, J., Rosenberg-Lee, M., Gazzaley, A., & Uncapher, M. R.

(2023). Better together: Novel methods for measuring and modeling development of executive function diversity while accounting for unity. *Frontiers in Human Neuroscience*, *17*. https://doi.org/10.3389/fnhum.2023.1195013

# Bayesian Distributional Models of Executive Functioning

Robert Kasumba[1], Zeyu Lu[2], Dom CP Marticorena[3], Mingyang Zhong[3], Paul Beggs[3], Anja Pahor[4], Geetha Ramani[5], Imani Goffney[6], Susanne M Jaeggi[7], Aaron R Seitz[7], Jacob R Gardner[8], Dennis L Barbour[1,2,3]

[1]Division of Computing and Data Science, Washington University, 1 Brookings Drive, St. Louis, MO 63130
[2]Department of Computer Science and Engineering, Washington University, 1 Brookings Drive, St. Louis, MO 63130
[3]Department of Biomedical Engineering, Washington University, 1 Brookings Drive, St. Louis, MO 63130
[4]Department of Psychology, University of Maribor, 2000 Maribor, Slovenia
[5]Department of Human Development and Quantitative Methodology, University of Maryland, College Park, MD 20742-5025
[6]Department of Teaching and Learning, Policy and Leadership, University of Maryland, College Park, MD 20742-5025
[7]Department of Psychology, Northeastern University, Boston, MA 02115
[8]Department of Computer and Information Science, University of Pennsylvania, Philadelphia, PA 19104

## Supplementary information for the manuscript with the same title

# 1. Supplementary analysis of model accuracy under matched observations

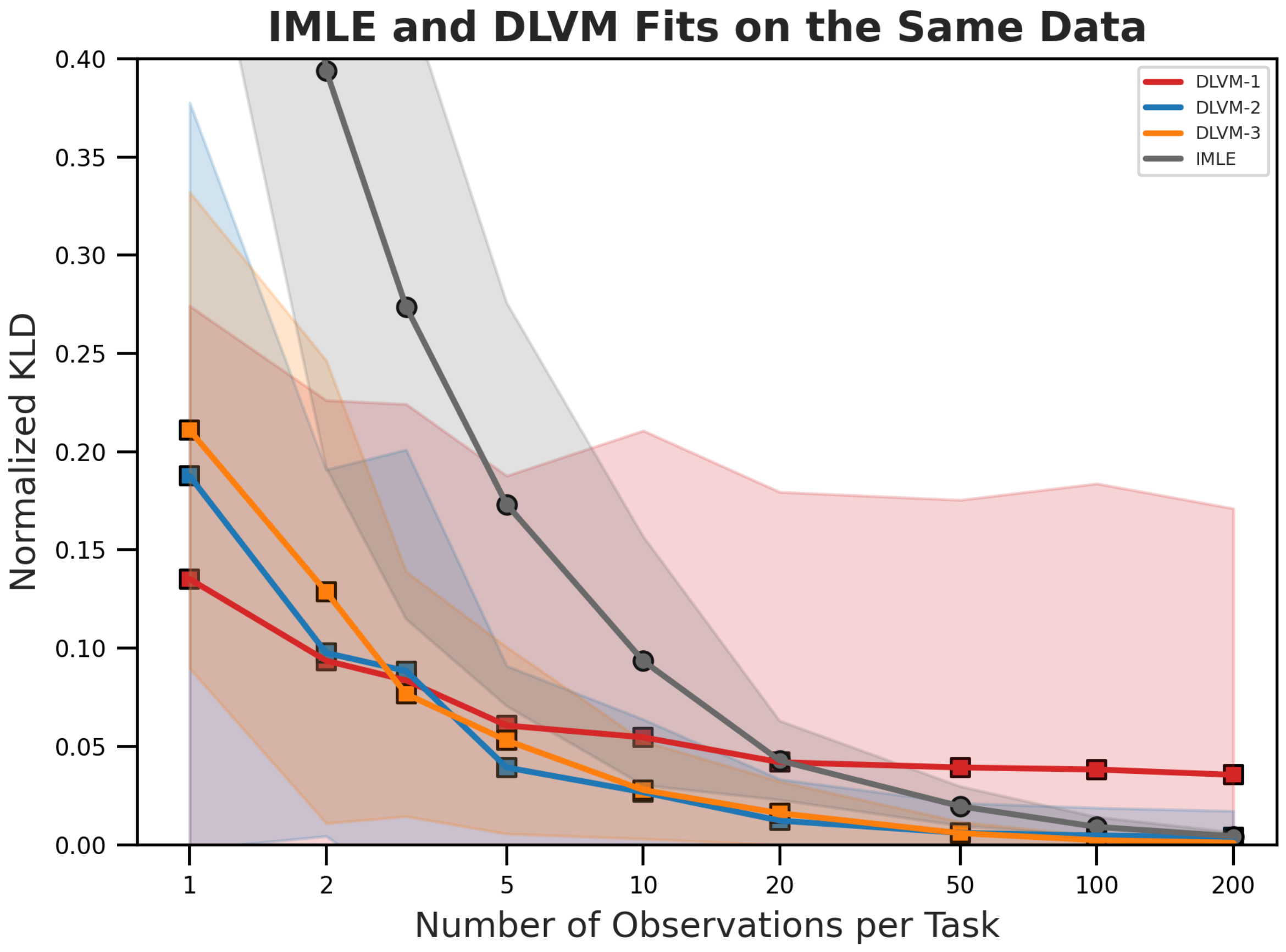

**Figure S1**: Mean ± standard deviation of DLVM (for three different model dimensionalities) and IMLE model accuracy ($n$ = 88 each) as the number of observations per task increases.

# 2. Supplementary analysis of DALE trajectories after 240 observations

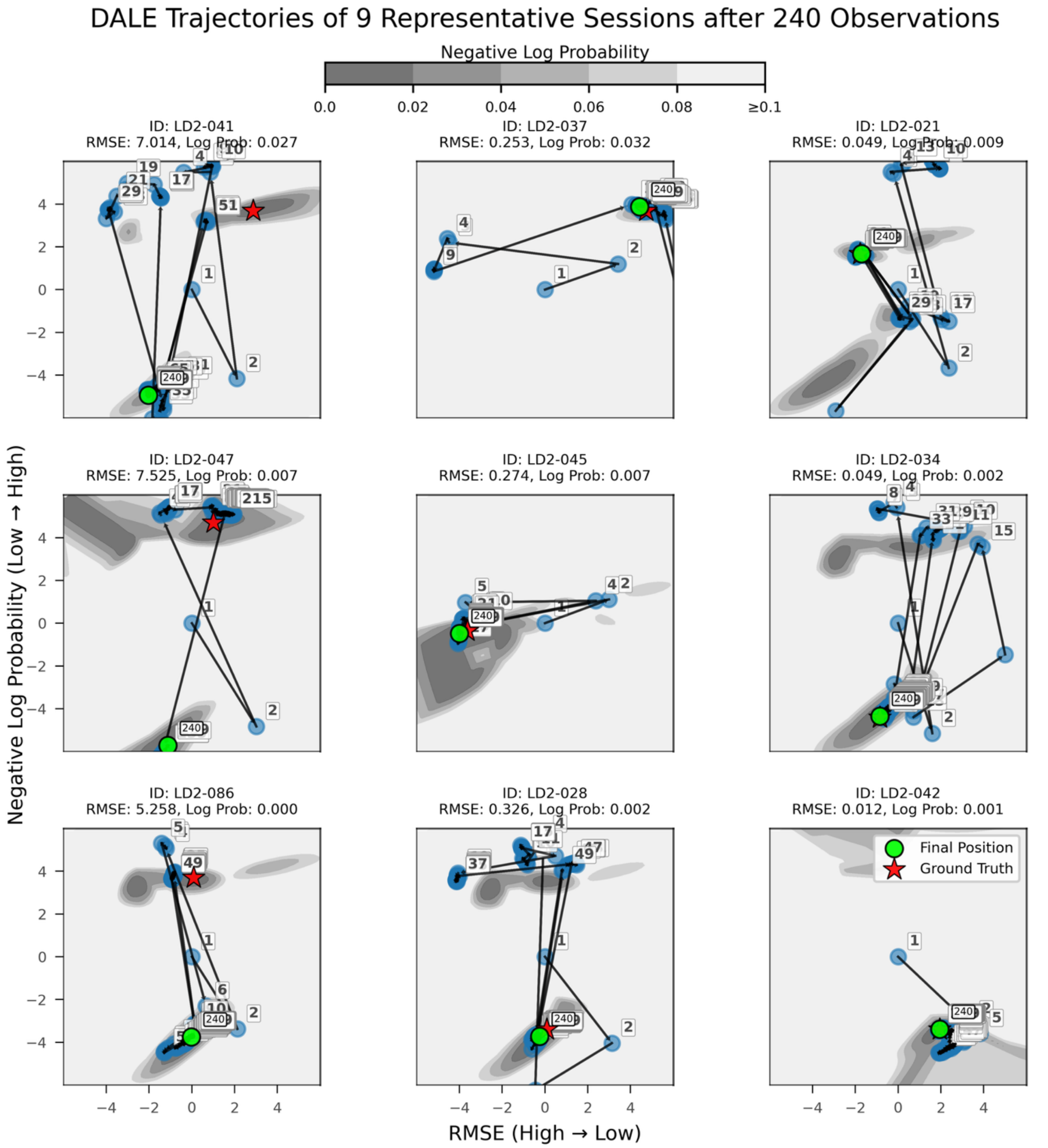

**Figure S2**: DLVM-2+DALE latent position updates as more data are collected for 9 representative simulated sessions based on the RMSE values between the ground truth position and the position estimated by DALE after accumulating 240 observations.

# 3. Supplementary analyses of latent dimensionality

## 3.1 Analysis of a One-Dimensional DLVM

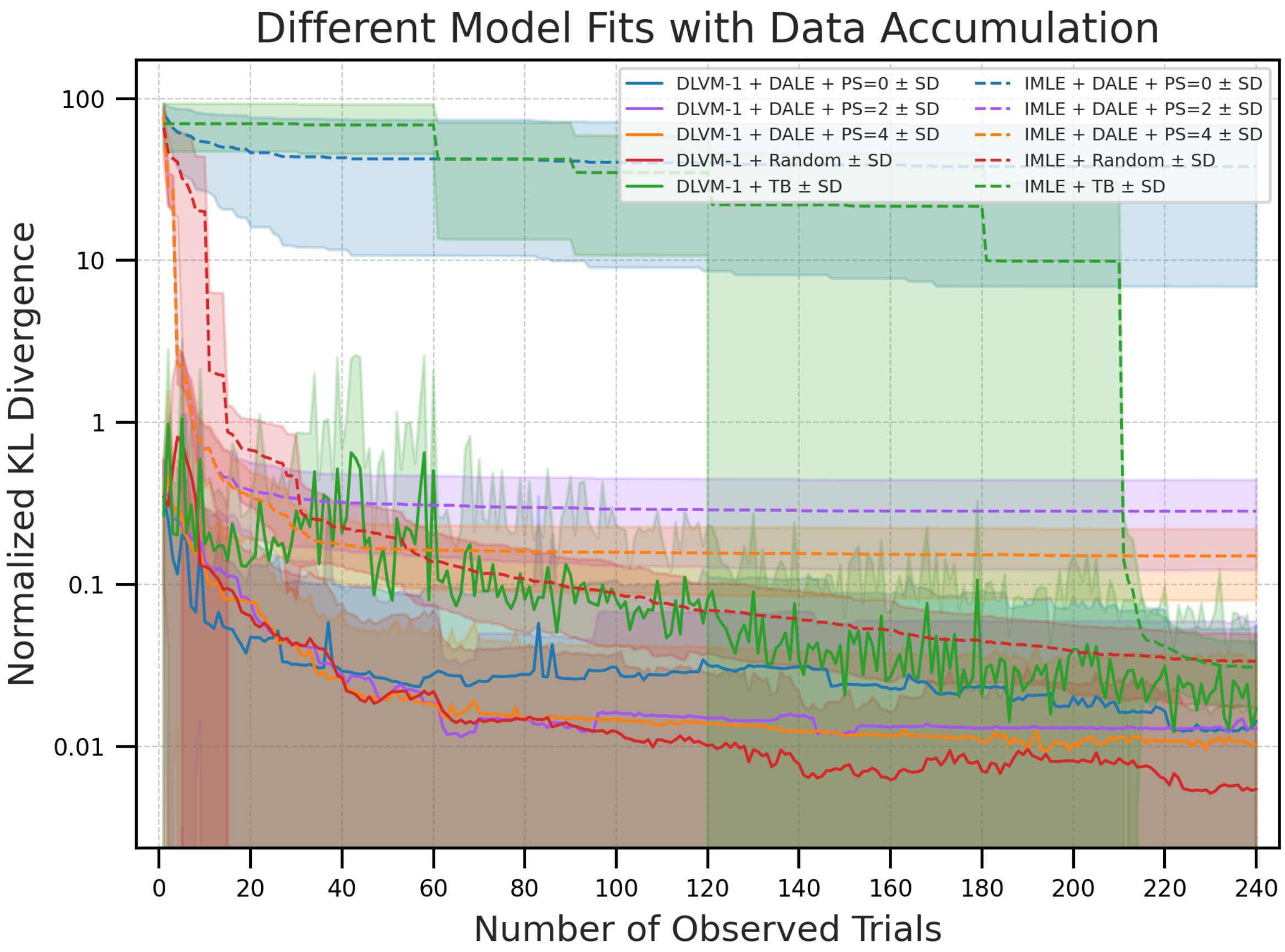

**Figure S3**: Performance of the different estimation methods and the sampling strategies for a 1D model. "PS=0" indicates that there was no primer sequence provided for DALE and thus the first sample was actively sampled, "PS=2" means that DALE was primed with 2 observations per task (16 total) and "PS=4", 4 observations per task (32 total) before active sampling started.

## 3.2 Analysis of a Two-Dimensional DLVM

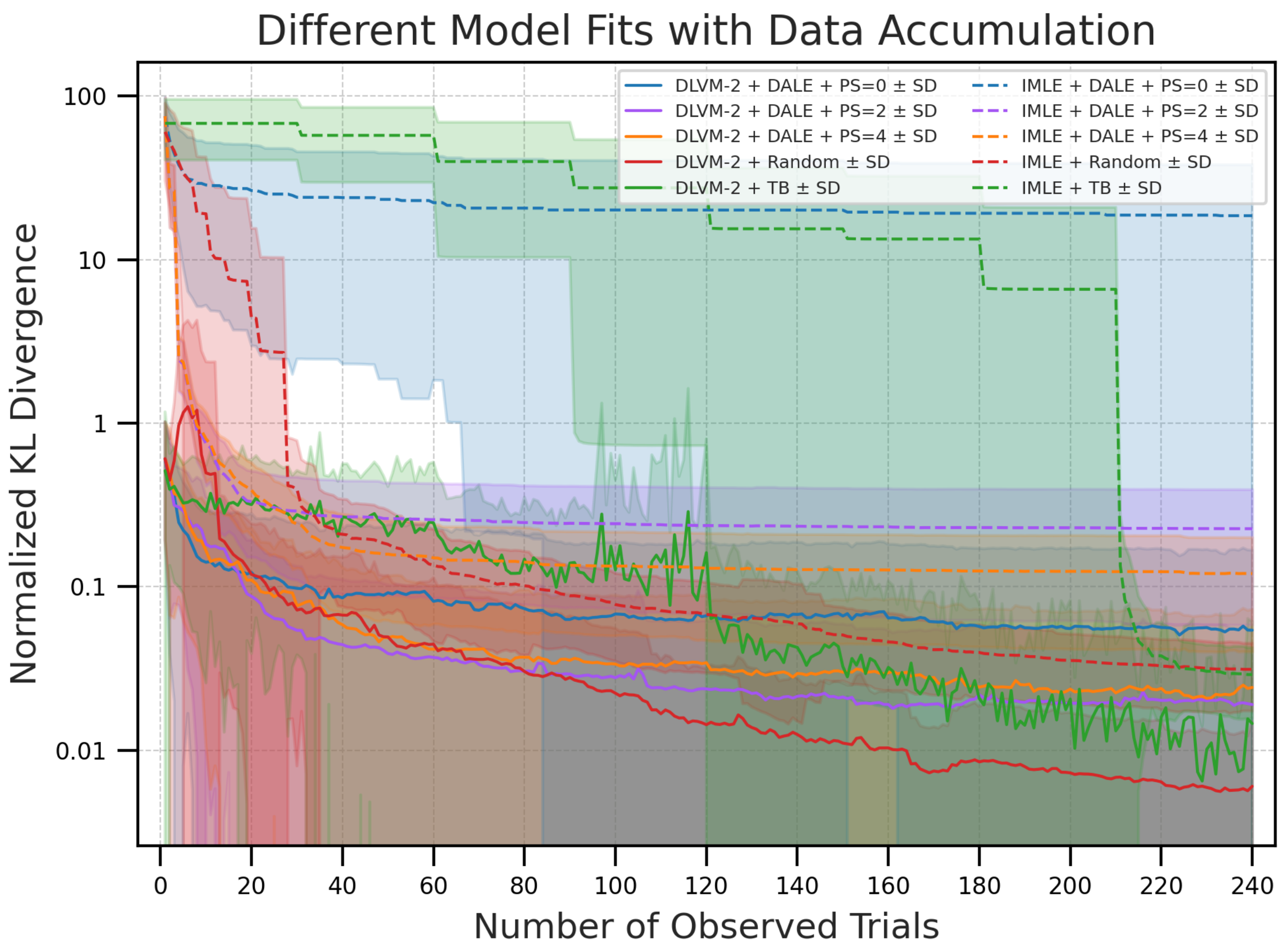

**Figure S4**: Performance of the different estimation methods and the sampling strategies for a 2D model. "PS=0" indicates that there was no primer sequence provided for DALE and thus the first sample was actively sampled, "PS=2" means that DALE was primed with 2 observations per task (16 total) and "PS=4", 4 observations per task (32 total) before active sampling started.

## 3.3 Analysis of a Three-Dimensional DLVM

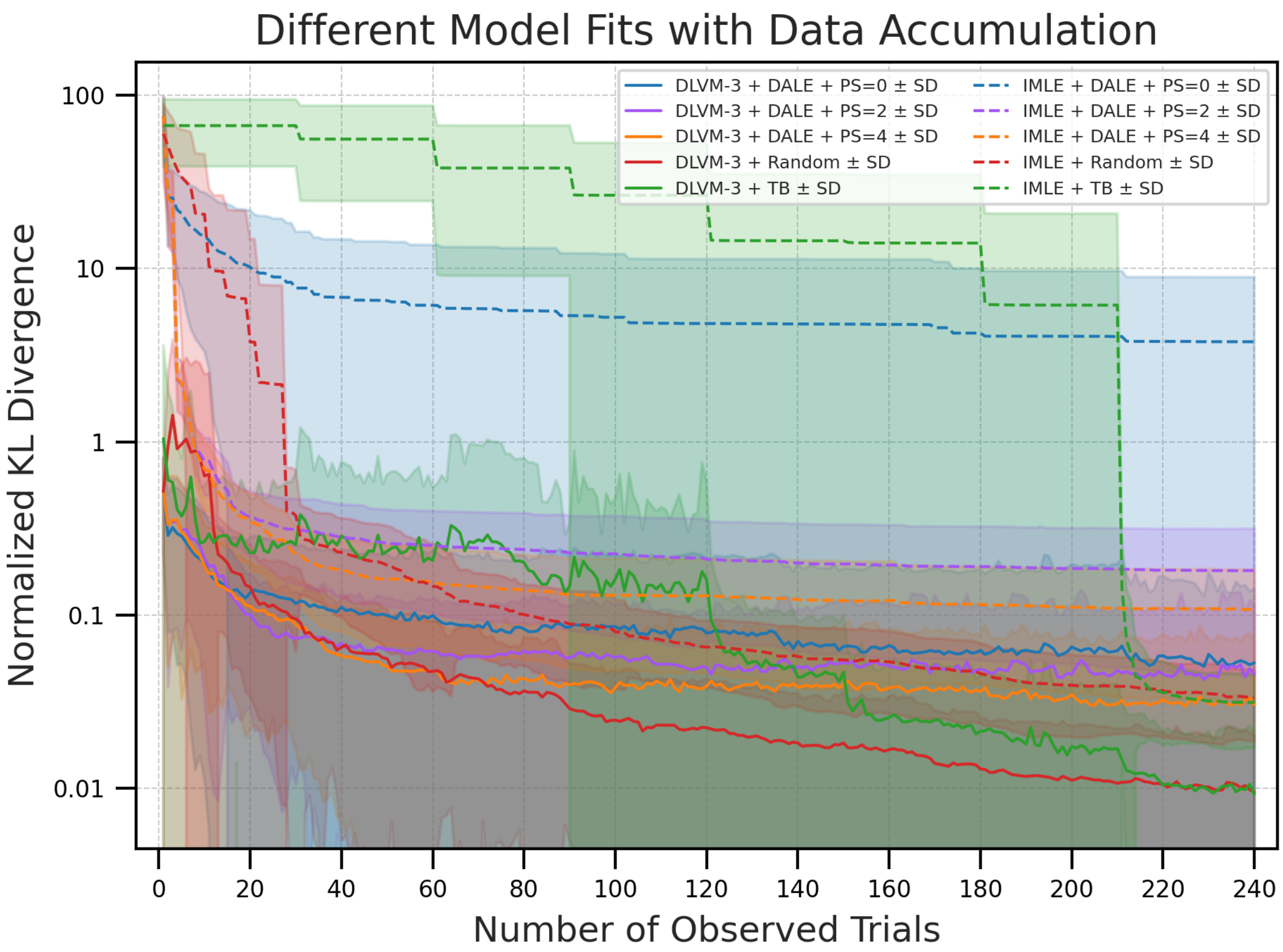

**Figure S5**: Performance of the different estimation methods and the sampling strategies for a 3D model. "PS=0" indicates that there was no primer sequence provided for DALE and thus the first sample was actively sampled, "PS=2" means that DALE was primed with 2 observations per task (16 total) and "PS=4", 4 observations per task (32 total) before active sampling started.

# 4. Supplementary analyses using real data (IMLE fitted) distributions as the generative process

This analysis evaluates whether the primary results depend on using a DLVM-based generative process. Because the target distributions in this analysis are IMLE-fitted distributions, IMLE is expected to perform well when sufficient observations are available. The key question is whether DLVM and DALE retain advantages under sparse observations, which we observe.

## 4.1 Matched data observation performance for IMLE and DLVM

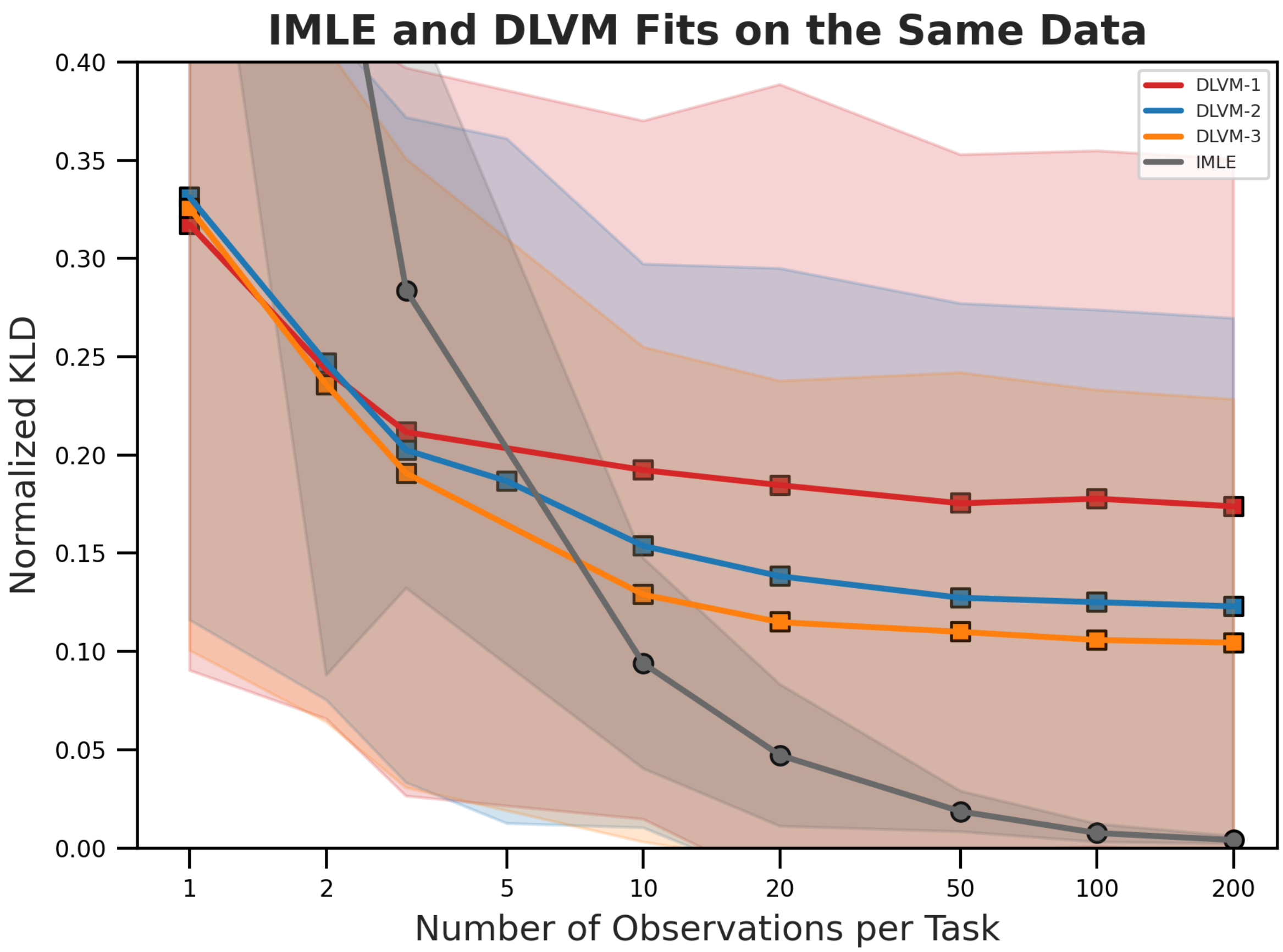

**Figure S6**: Model Accuracy DLVM (for three different model dimensionalities) and IMLE when the observations are sampled from **IMLE-fitted distributions on real human data** ($n$ = 88 each). The models aim to recover those IMLE-aligned distributional parameters.

## 4.2 One-dimensional DALE model

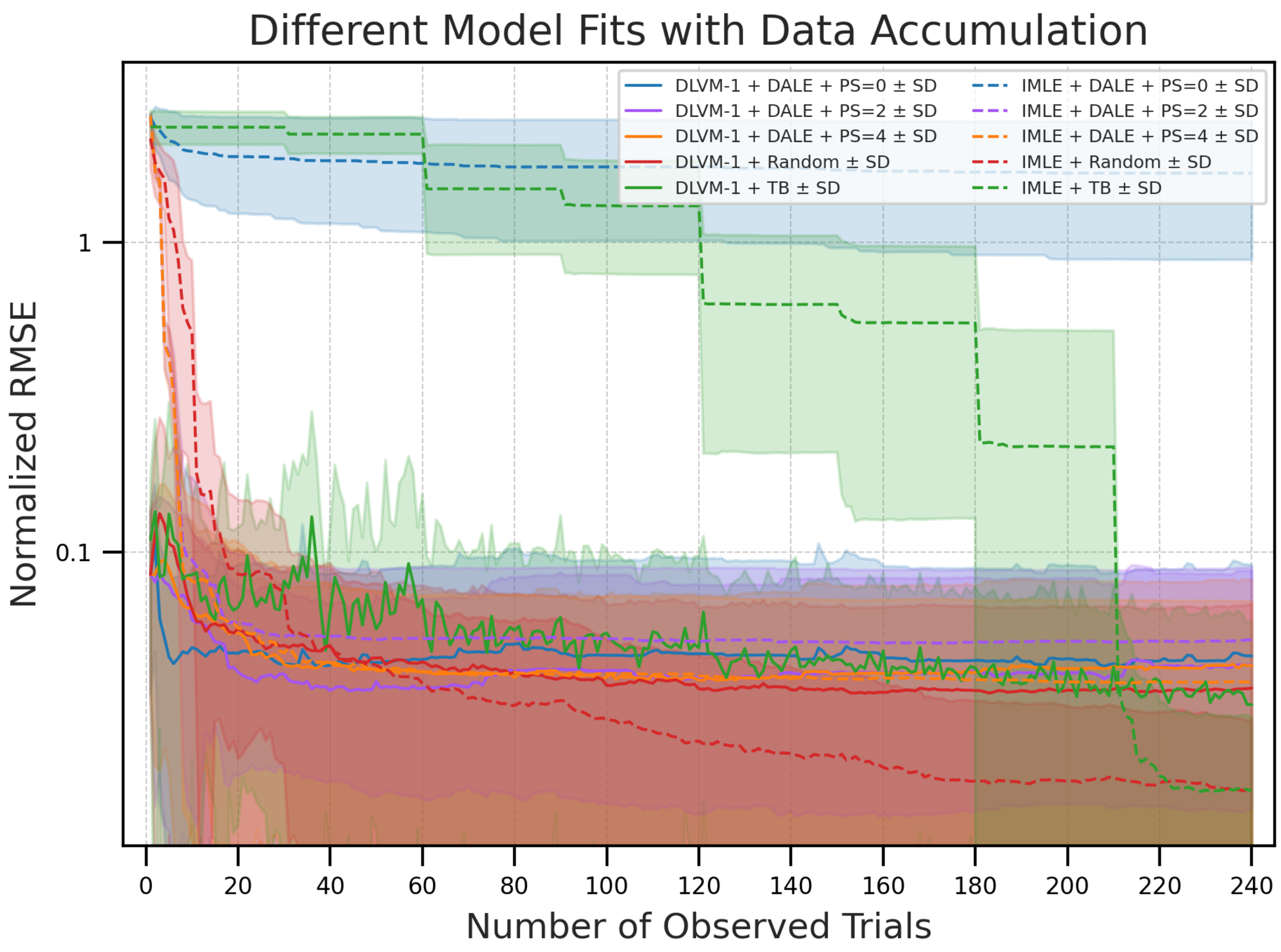

**Figure S7**: Performance of the different estimation methods and the sampling strategies for a 1D model when the observations are sampled from IMLE-fitted distributions on real human data.

## 4.3 Two-dimensional DALE model

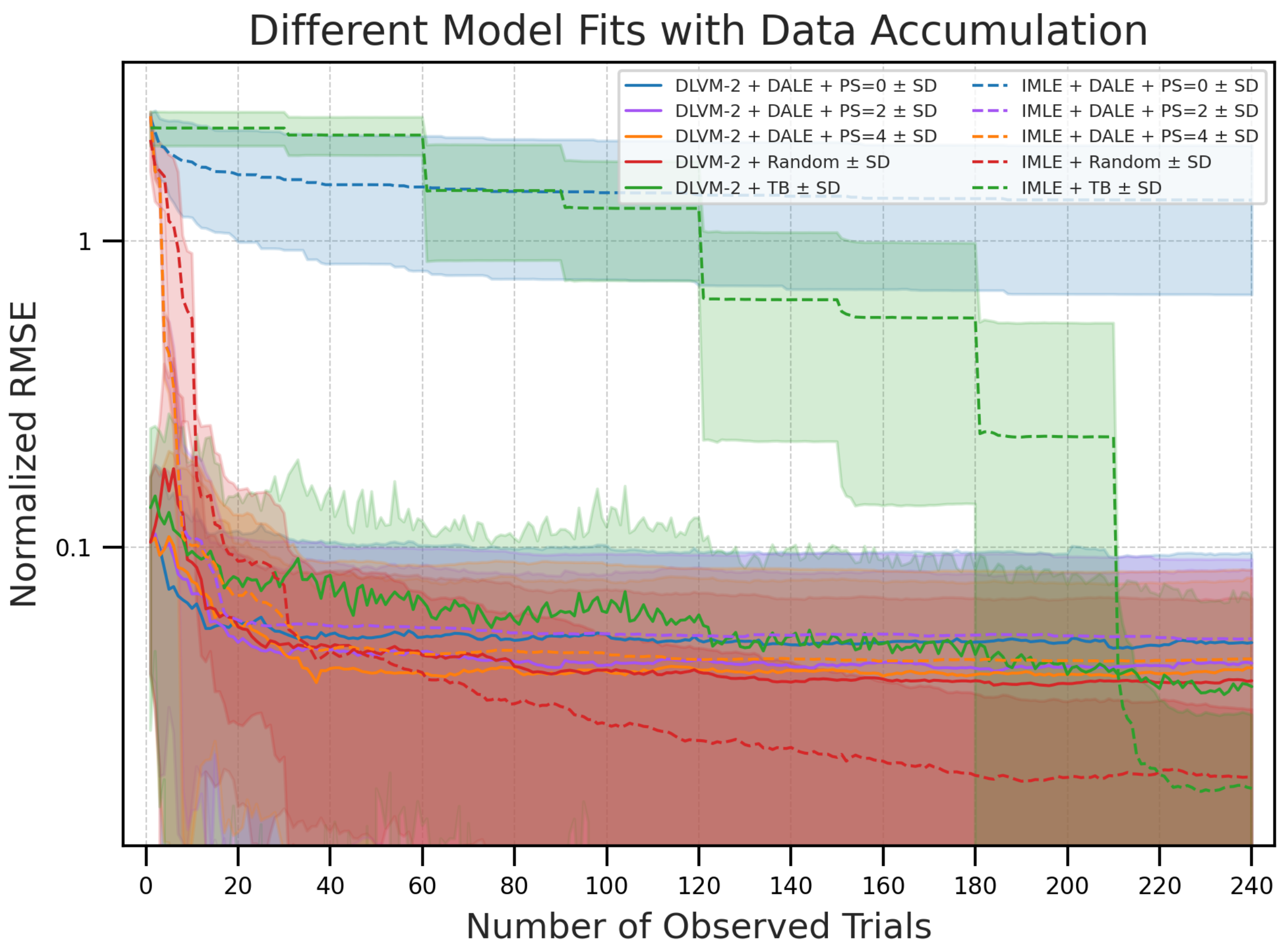

**Figure S8**: Performance of the different estimation methods and the sampling strategies for a 2D model when the observations are sampled from IMLE-fitted distributions on real human data.

## 4.4 Three-dimensional DALE model

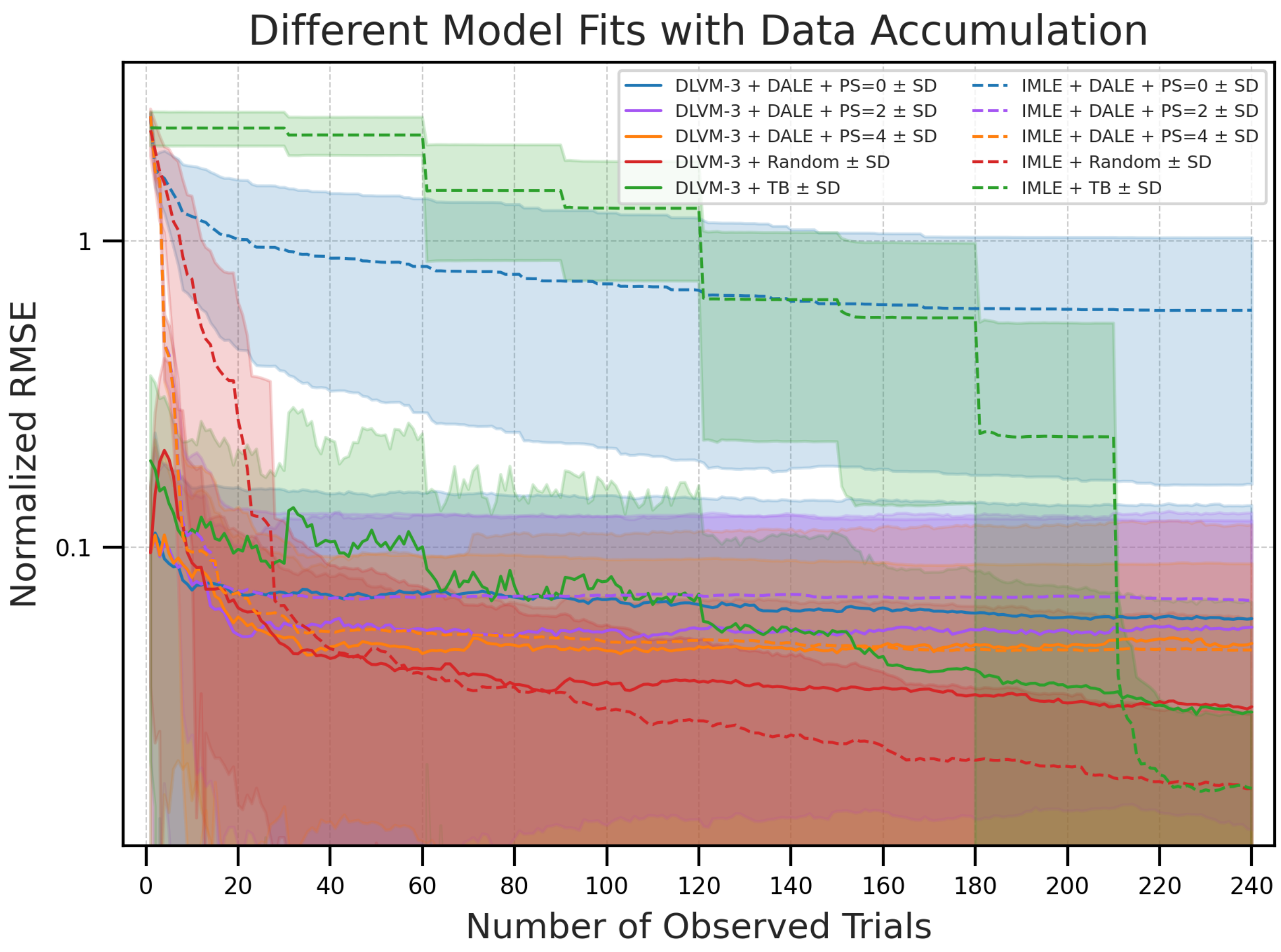

**Figure S9**: Performance of the different estimation methods and the sampling strategies for a 3D model when the observations are sampled from IMLE-fitted distributions on real human data.

# 5. Robustness analysis with outliers retained

The primary analysis used DLVM models trained with outlier sessions removed, retaining a total of $n$ = 88 sessions. We retrained a 2-dimensional DLVM model without removing outlier sessions, a total of $n$ = 101 sessions. Here we present this supplementary analysis. Figure S10 shows the learned latent space as well as the sampled synthetic individuals, and Figure S11 shows DALE performance. We evaluated DALE using 101 synthetic individuals using uniform sampling as in the primary analysis.

## 5.1 Learned DLVM latent space

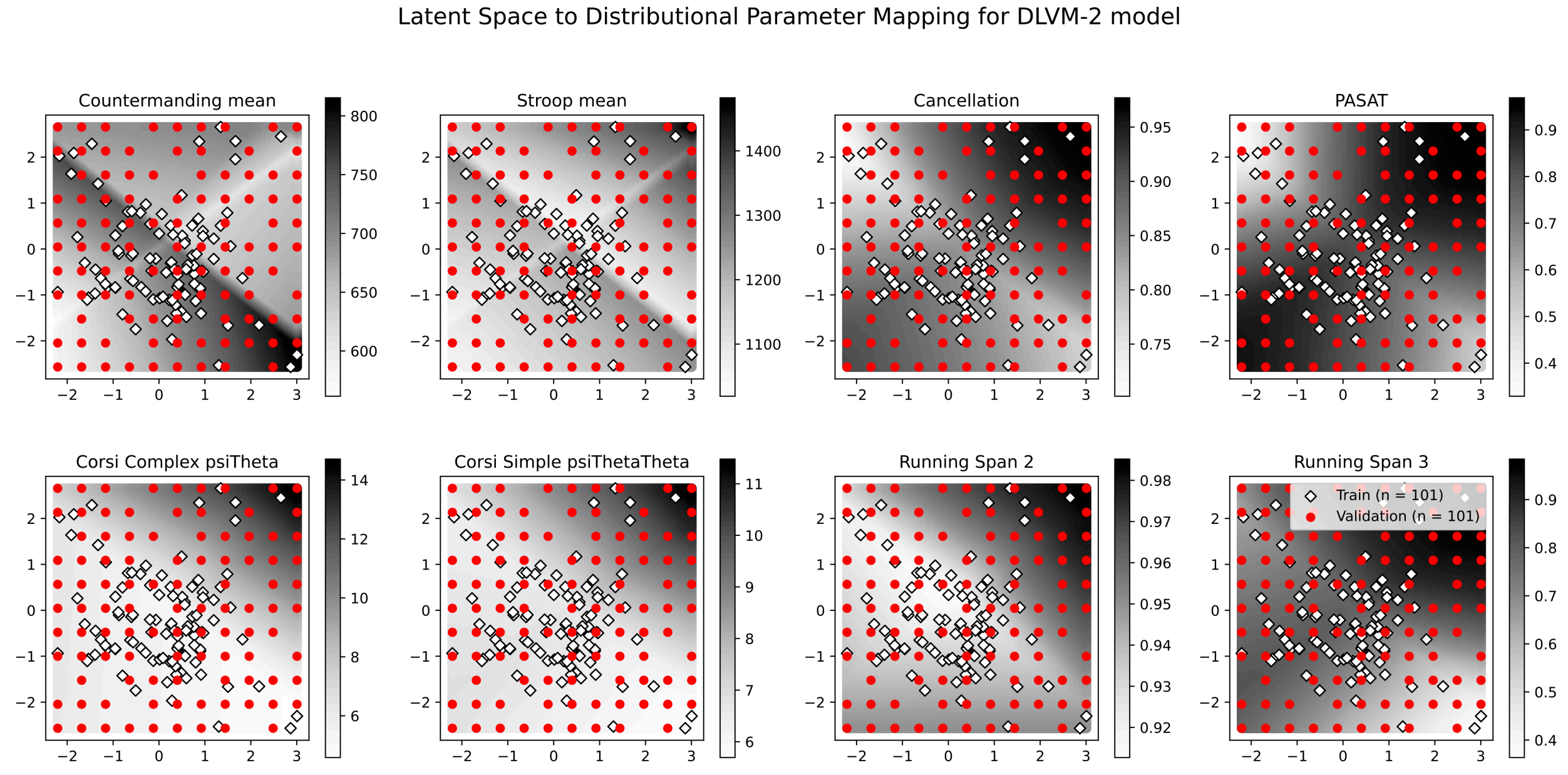

**Figure S10**: DLVM-2 latent-space-to-parameter mappings when outliers are retained. Compared with the primary DLVM-2 mapping shown in Figure 2, the expanded dataset produces visibly more locally irregular parameter surfaces while preserving the broad cross-task organization of the latent space. The additional sessions introduce sharper transitions and stronger nonlinear curvature in several mappings, most notably for Countermanding mean, Stroop mean, PASAT, and Running Span 3.

## 5.2 Analysis of Two-Dimensional DALE model

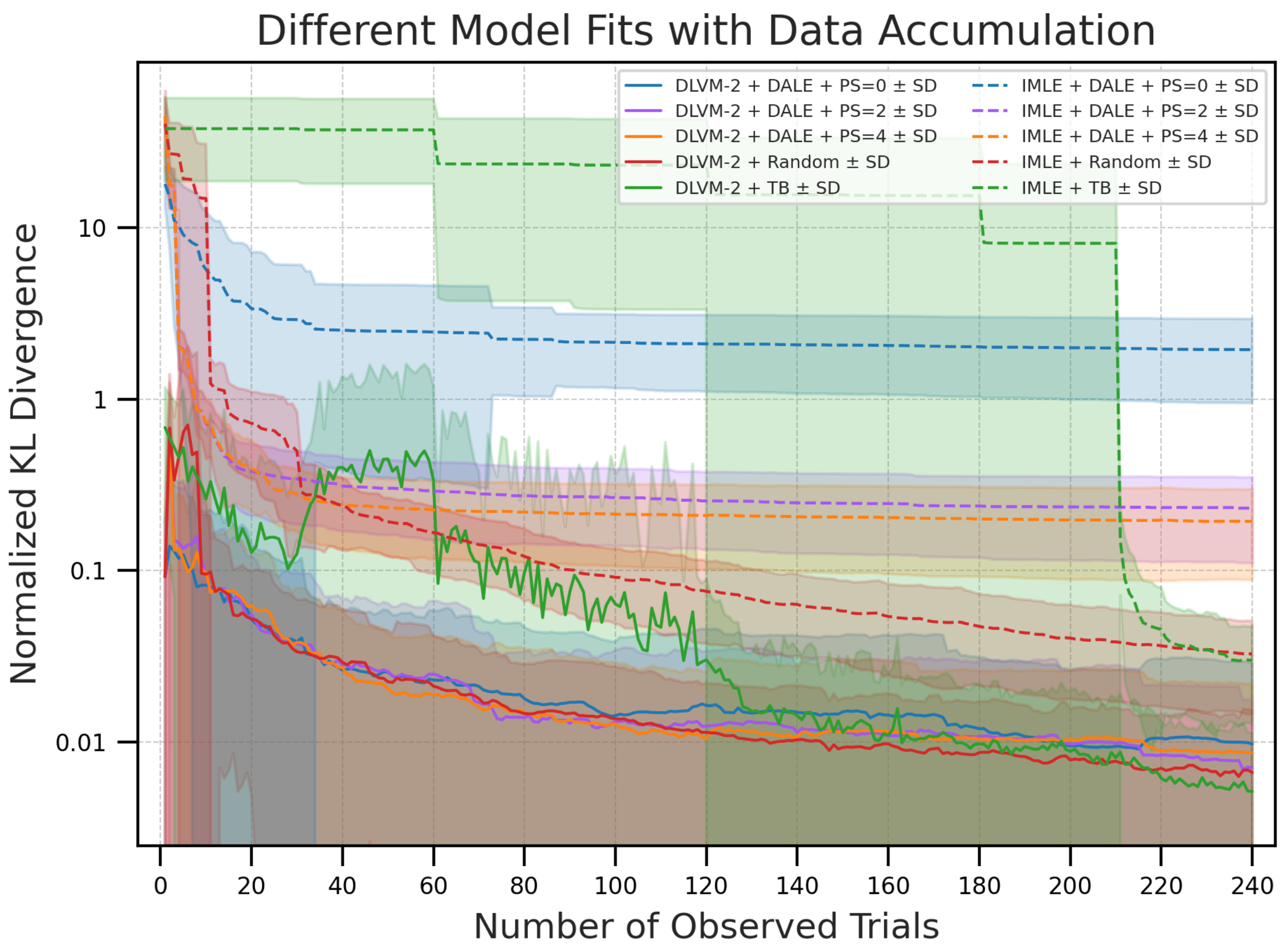

**Figure S11**: Performance of the different estimation methods and the sampling strategies for a 2D model when generative space was learned with outliers included. The general trend is consistent with the primary analysis, with DALE showing a slight advantage over random sampling within the first 100 observations. The advantage is less pronounced, however, compared to when the latent space is smoother.